\documentclass{article}

\usepackage{arxiv}

\usepackage[utf8]{inputenc} 
\usepackage[T1]{fontenc}    
\usepackage{hyperref}       
\usepackage{url}            
\usepackage{booktabs}       
\usepackage{amsfonts}       
\usepackage{nicefrac}       
\usepackage{microtype}      
\usepackage{subcaption}
\usepackage{lipsum}
\usepackage{amsmath}
\usepackage{multirow}
\usepackage{graphicx}
\usepackage[numbers,sort&compress]{natbib}
\graphicspath{ {./images/} }

\title{AdaPolicy: Action-Driven Anticipatory Robot Policies via Next-Chunk Prediction}

\author{%
\begin{tabular}{c}
Fan Yang\textsuperscript{1} \quad
Yuting Su\textsuperscript{1} \quad
Chen Zhao\textsuperscript{2} \quad
Fugui Fan\textsuperscript{1} \quad
Peiguang Jing\textsuperscript{1} \\[0.45em]
\textsuperscript{1}Tianjin University \quad
\textsuperscript{2}University of Bologna \\
[3pt]
\texttt{eeyf@tju.edu.cn}
\end{tabular}
}

\begin{document}
\pagestyle{plain}

\maketitle
\begin{abstract}

Endowing robot policies with anticipatory capabilities has proven effective in improving control performance.
However, existing methods predominantly rely on explicit supervision derived from future observations, in either pixel or latent space, at the cost of additional architectural and computational complexity.
Departing from explicit future supervision, we ask whether such foresight can be elicited directly from the action signal.
To this end, we propose \textbf{AdaPolicy}, an \textbf{A}ction-\textbf{d}riven \textbf{a}nticipatory policy framework that shapes future-aware representations via cascaded next-chunk prediction.
Specifically, an Online Action Head jointly produces the current action chunk and an evolved latent state. 
Building on this state, a cascaded Unrolled Action Head forecasts the next chunk, and the resulting loss flows back through the evolved state to instill implicit foresight into the online policy. 
Importantly, this training-only unroll requires no future-observation encoding, keeping AdaPolicy lightweight to train with zero inference-time overhead.
Experiments on MetaWorld, RoboTwin 2.0, LIBERO, and real-world deployments demonstrate consistent gains across diverse settings, including up to $14$ percentage points on millimeter-precision tasks such as earphone assembly and shoelace threading.
More details and visualizations are available at \href{https://teee111.github.io/AdaPolicy/}{https://teee111.github.io/AdaPolicy/}.

\end{abstract}


\section{Introduction}

As vision-language-action (VLA) models continue to advance generalist robot 
control~\cite{openvla,community2026starvla,vla_adapter}, a growing body of work shows that robot policies 
benefit from looking ahead.
In pursuit of this anticipatory capability, recent efforts have incorporated future-oriented supervision into policy learning~\cite{jiang2026rethinking,intelligence2026pi,sun2026vla,lyu2026lda,bi2025motus}.
Such supervision takes diverse forms, including future images or videos~\cite{ni2026gigaworld,kim2026cosmos,efficientwam,black2024zero}, features extracted from future observations~\cite{lewm,flare,intact,gu2026right,lewam}, and structured targets~\cite{scaleflow,robopoint,xu2025a0,lee2026molmoact}.
Despite their diversity, these objectives share a common reliance on targets derived from future observations. 
In this work, we explore an alternative: eliciting temporal foresight directly from the action signal, without explicit supervision from future observations.

\begin{figure*}[t]
\centering
\includegraphics[width=\textwidth, trim=5 325 350 28,clip]{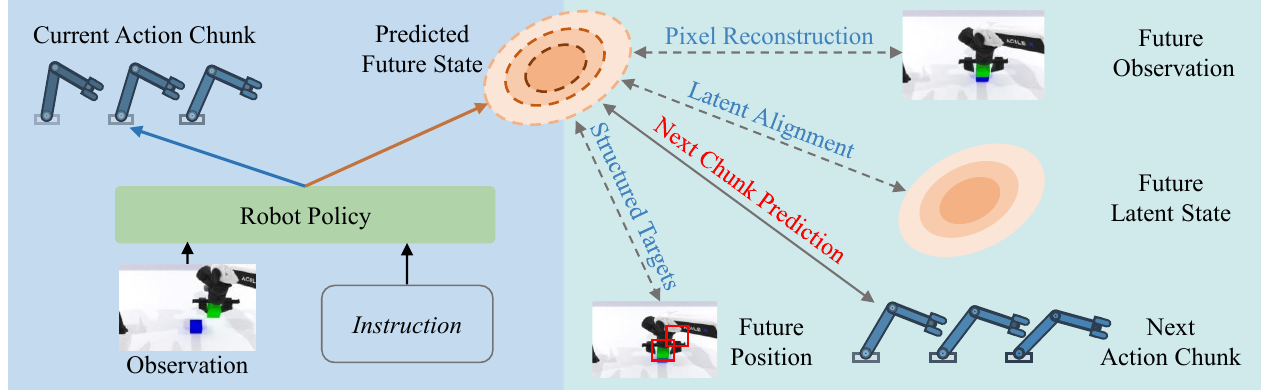} 
\caption{\textbf{Comparison of future-supervision signals for robot policies.}
Existing methods supervise the predicted future state with explicit targets derived from 
future observations, including pixel reconstruction, latent alignment, and structured 
geometric targets (e.g., future object positions).
In contrast, \textbf{AdaPolicy} uses the next action chunk as a control-aligned supervision signal, requiring no explicit future-observation targets.}
\label{fig:IdeaPic}
\end{figure*}

Existing methods typically instantiate explicit future supervision in three main ways (Figure~\ref{fig:IdeaPic}): reconstructing future pixels, aligning with future latent features, and predicting structured geometric targets.
(1) World-Action Models (WAMs) predominantly learn by reconstructing future images or 
videos~\cite{pai2025mimic,shen2025videovla,hu2024video,bi2025motus,arwam,ni2026gigaworld,kim2026cosmos,efficientwam,black2024zero}, 
but pixel-space targets devote substantial capacity and computation to appearance 
details peripheral to control.
(2) Latent-space methods instead operate on features of future frames to avoid raw-pixel reconstruction. 
Some approaches directly align policy states with future embeddings~\cite{flare,lilawam,gu2026right,last0}, relying on frozen extractors~\cite{flare} or dedicated anti-collapse losses~\cite{beingbeyond2026beingh07,intact}. 
Others adopt a latent-action paradigm~\cite{lapa,moto,lawam}, first learning latent actions and a latent world model from observation pairs before distilling them into the policy in a second stage.
(3) Structured geometric targets, such as object bounding boxes and point trajectories~\cite{scaleflow,robopoint,xu2025a0,lee2026molmoact}, are more selective, 
but rely on constructed annotations and may discard control-relevant information.
Across all three forms, the supervision signal is derived from future observations.
However, such signals are only indirectly aligned with action 
generation: policies are ultimately designed to produce 
actionable control, and faithfully reconstructing future details 
is unnecessary for this purpose~\cite{grimm2020value,lewam}.

A natural question therefore arises: \emph{can supervision from subsequent actions shape anticipatory latent dynamics without explicit supervision from future observations? }
In standard VLA policies, intermediate state representations are only supervised by the current action chunk; without constraints toward the future, we observe that the output state tokens lack the incentive to progress forward in time and tend to degenerate into a near-identity copy of the current input condition.
To convert subsequent-action supervision into an implicit driver for such foresight, we introduce a simple formulation: the policy outputs an evolved latent state alongside the current action chunk, and conditions the prediction of the next action chunk exclusively on this evolved state. 
Crucially, to reduce the next-chunk prediction error, the evolved state can no longer merely copy the current observation features; instead, it is forced to integrate the ongoing action trajectory and encode information useful for determining subsequent actions, thereby implicitly modeling the future state. 
This formulation offers two key advantages: first, it elicits anticipatory state transitions without reconstructing future observations or aligning with features extracted from them; second, because the supervision arrives in the same space as the policy's own output, the evolved state is optimized purely for its utility in control rather than spending capacity on control-irrelevant visual details.

Building on this insight, we propose \textbf{AdaPolicy} (\textbf{A}ction-\textbf{d}riven \textbf{a}nticipatory Policy), a framework that shapes future-aware representations in imitation learning via cascaded next-chunk prediction. 
A Context Adapter uses learnable queries to compress features from a pretrained backbone into a compact condition state. 
The Online Action Head, implemented as a Diffusion Transformer (DiT)~\cite{peebles2023scalable}, takes this condition state as input and outputs an evolved state alongside the current action chunk. 
Conditioned solely on this evolved state, a cascaded Unrolled Action Head predicts the next action chunk. 
To prevent the two heads from trivially co-adapting and to stabilize the next-chunk supervision, the unrolled head is parameterized as a gradient-decoupled exponential moving average (EMA) copy of the online head~\cite{he2020momentum, grill2020bootstrap}. 
Crucially, this latent unroll operates strictly as a training-time shaping mechanism, making AdaPolicy plug-and-play with zero inference-time overhead. 
We evaluate AdaPolicy across MetaWorld, RoboTwin 2.0, LIBERO, and six real-world manipulation tasks, demonstrating its broad applicability across diverse environments as well as its effectiveness on challenging millimeter-precision tasks such as earphone assembly and shoelace threading.

Our main contributions are summarized as follows:

\begin{itemize}
\item \textbf{Action-Driven Anticipatory Policy Learning.} We propose AdaPolicy, an anticipatory policy framework for chunk-based imitation learning that converts subsequent-action supervision into an implicit driver for temporal foresight via cascaded next-chunk prediction. By conditioning an EMA-updated unrolled head solely on the evolved latent state, our method prevents the state representation from trivially copying the current input and elicits anticipatory latent dynamics with zero inference-time overhead.

\item \textbf{Evaluation Across Diverse Settings.}
Experiments on MetaWorld, RoboTwin 2.0, LIBERO, and real-world tasks requiring millimeter-level precision demonstrate the effectiveness of our approach across diverse settings.

\item \textbf{Representation Analysis.}
Probing and latent-space planning analyses confirm our central 
premise: next-chunk supervision shapes the policy's internal 
representations to be future-aware, without targets derived 
from future observations.
Concretely, future object positions become more decodable from 
these representations, and the representations better support 
planning toward future goals.
\end{itemize}

\section{Related Work}
\label{sec:related}

\subsection{Explicit Future Supervision in Robot Policies}

A prominent line of work equips robot policies with explicit supervision derived 
from future observations, differing primarily in \emph{what} the future target is.
A widely adopted choice is pixel space, where video prediction supplies the 
supervisory signal: predictive representations can be extracted from pretrained 
video generation models~\cite{hu2024video,kim2026cosmos}, or future video and 
actions can be modeled jointly within a single generative 
process~\cite{bi2025motus,li2026causal,shen2025videovla,pai2025mimic,cen2025worldvla,univla,feng2025vidar,dreamzero}.
Pixel-level targets, however, devote substantial capacity and computation to 
appearance details peripheral to control.

A second line instead derives targets in latent space.
FLARE~\cite{flare} aligns policy features with embeddings of future observations, 
WoG~\cite{wog} distills compact future representations through a two-stage pipeline, 
Being-H0.7~\cite{beingbeyond2026beingh07} relies on dedicated anti-collapse regularizers, 
and ImageWAM~\cite{zhang2026imagewam} repurposes intermediate features of image-editing 
models.
Another prominent latent-space paradigm centers on latent actions~\cite{lapa,moto,lawam}, which infer visual transition tokens between current and future observations and use them as an intermediate bridge for downstream policy learning.
Because these methods must first train a tokenizer~\cite{lapa} or inverse-dynamics model~\cite{lawam} to extract latent actions before supervising or distilling into the policy, they inherently rely on multi-stage training pipelines.
While bypassing pixel rendering, both direct alignment and latent-action approaches still derive their supervisory signals from future visual observations, whereas our method operates in a single end-to-end stage directly within the action stream without auxiliary tokenizers or multi-stage distillation.

A third line supervises structured geometric targets, such as waypoints, object 
bounding boxes, keypoints, point tracks, or optical 
flow~\cite{huang2025roboground,ma2025glover,ji2025robobrain,reconvla,xu2025a0,zhao2026towards,sundaresan2023kite,pointflow,fan2026aim,scaleflow,robopoint,lee2026molmoact,bharadhwaj2024track2act,vecerik2024robotap,seita2023toolflownet,unipi}.
Such targets are more selective, but require hand-designed annotations and may omit control-relevant information.

More recent work questions the necessity of explicit future generation itself: 
UVA~\cite{uva} and UWM~\cite{uwm} bypass video decoding at inference while retaining video-action 
co-training.
Fast-WAM~\cite{yuan2026fast} further shows that much of the benefit stems from 
training-time video modeling rather than test-time 
imagination.
Across all these designs, however, the supervisory signal still originates from 
future observations, whether in the form of pixels, latents, or geometric targets.
\textbf{AdaPolicy} takes this line of questioning one step further, demonstrating that cascaded next-chunk prediction alone can endow policies with temporal foresight, without any explicit supervision from future observations.

\subsection{Learning Anticipatory Representations via Latent Unrolling}

The idea of shaping internal representations without reconstructing future observations has roots in model-based reinforcement learning, where latent dynamics models are learned and unrolled over multiple steps for prediction or planning.
Value Prediction Networks~\cite{vpn} and the Predictron~\cite{silver2017predictron} first trained abstract states to predict future values rather than future observations.
MuZero~\cite{muzero} advanced this principle: its latent states are unrolled over hypothetical actions and shaped by reward, value, and policy targets, with no requirement to reconstruct observations or match true environment states.
Subsequent systems reintroduced observation-derived guidance: EfficientZero~\cite{efficientzero} adds latent consistency against encoded future observations, TD-MPC~\cite{tdmpc} and TD-MPC2~\cite{tdmpc2} combine value learning with latent temporal consistency, and the Dreamer~\cite{dreamerv3} family trains latent dynamics with explicit observation reconstruction.

Among these methods, MuZero~\cite{muzero} shares a similar philosophical motivation: shaping 
latent states with targets derived from the decision-making 
process rather than future observations, as formalized by the value-equivalence 
framework~\cite{grimm2020value,grimm2021proper}.
AdaPolicy adapts this philosophy to reward-free, chunk-based imitation 
learning by requiring the evolved state to support the generation of subsequent action chunks.
However, instantiating this in generative action policies differs fundamentally from MuZero in both mechanism and optimization dynamics.
While MuZero unrolls single-step transitions and backpropagates gradients through a shared model, modern chunk-based policies generate continuous trajectories via iterative denoising, where output state tokens lack explicit future constraints and tend to degenerate into a near-identity copy of the input condition state.
To turn the next-chunk prediction loss into a state-transition constraint, AdaPolicy decouples the denoising timesteps across stages and parameterizes the unrolled head as an EMA-updated copy excluded from direct gradient updates. 
Because the model cannot reduce the next-chunk error by directly updating the unrolled head, the optimization pressure is routed exclusively through the evolved state, preventing it from merely copying the current input and forcing it to anticipate future control-relevant dynamics.

In the imitation learning setting, Next Forcing~\cite{nextforcing} also adopts a multi-chunk unroll design, but differs fundamentally in its supervision space and mechanism.
Specifically, Next Forcing unrolls in an observation-derived latent space by autoregressively predicting future video latents.
This approach still requires encoding future visual streams during training and expends capacity on modeling appearance details peripheral to control.
In contrast, AdaPolicy unrolls at the granularity of action chunks purely within the action stream.
This design avoids the computational overhead of future pixel reconstruction or feature alignment, while providing a directly control-aligned supervisory signal through the next action chunk.

\begin{figure}[t]
\centering
\includegraphics[width=\textwidth, trim=9 315 441 9,clip]{ 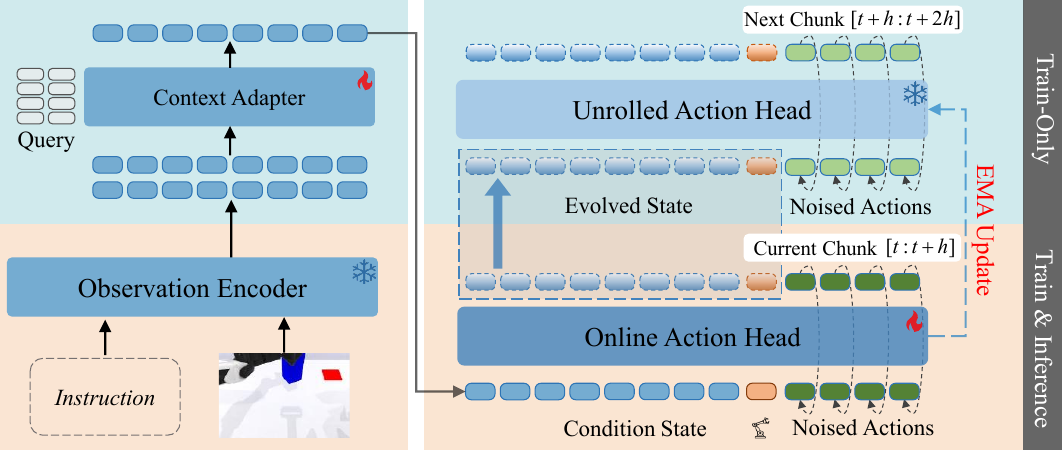} 
\caption{
\textbf{Architectural overview.} 
Left: The Context Adapter compresses pretrained features into compact conditioning tokens.
Right: The Online Action Head jointly predicts the current action chunk and an evolved latent state, which conditions the Unrolled Action Head to predict the next chunk.
During training, the next-chunk loss propagates through the evolved state to the online pathway; the cascaded step is omitted at inference time.
}
\label{fig:Framework}
\end{figure}

\section{Method}
\label{sec:method}

We present \textbf{AdaPolicy}, an action-driven anticipatory policy that shapes future-aware representations via cascaded next-chunk prediction, without any explicit targets derived from future observations.
Figure~\ref{fig:Framework} illustrates the overall architecture, which consists of a Context Adapter, an Online Action Head, and a training-time Unrolled Action Head.
Rather than supervising the evolved latent state with explicit future-observation targets, AdaPolicy shapes it implicitly through subsequent-action supervision.

\subsection{Problem Setup and Context Encoding}
\label{sec:context}

Let $o_t$, $l$, and $p_t$ denote the observation, task instruction, and proprioceptive state at time $t$, respectively.
We consider two consecutive action chunks of length $h$:
$\mathbf{A}_t^{(0)}=\mathbf{a}_{t:t+h}$ and
$\mathbf{A}_t^{(1)}=\mathbf{a}_{t+h:t+2h}$.
A standard chunk-based policy is trained to generate $\mathbf{A}_t^{(0)}$ from the current condition.
While the current-chunk objective shapes intermediate representations, it provides no explicit incentive for them to progress forward in time or encode temporal foresight.

By decoupling context compression from feature extraction, AdaPolicy is compatible with both pretrained visual and vision-language backbones~\cite{bai2025qwen3,simeoni2025dinov3}.
Given the backbone features of the current observation (and the instruction, when available), the Context Adapter $C_\psi$ compresses them with a set of learnable queries into compact conditioning tokens.
The proprioceptive state $p_t$ is separately projected by a linear layer and concatenated with these tokens along the token dimension; we denote the resulting conditioning state tokens as $\mathbf{c}_t$.

\subsection{Cascaded Next-Chunk Prediction}
\label{sec:cascade}

The Online Action Head $H_\theta$ adopts a DiT~\cite{peebles2023scalable} to enable the joint prediction of actions and the 
evolved latent state within a single module.
All tokens interact through bidirectional self-attention, allowing the head to 
produce the current-chunk velocity field and the evolved latent state in a single forward pass:
\begin{equation}
    \bigl(\hat{\mathbf{v}}_t^{(0)},\hat{\mathbf{z}}_{t+h}\bigr)
    =
    H_\theta\bigl(\mathbf{c}_t,\mathbf{x}_{\tau_0}^{(0)},\tau_0\bigr).
    \label{eq:online_head}
\end{equation}
Here, $\tau_0$ is the flow-matching time for the current chunk, and $\mathbf{x}_{\tau_0}^{(0)}$ denotes its noisy action tokens.
The updated readout corresponding to the conditioning state tokens $\mathbf{c}_t$ constitutes the evolved latent state $\hat{\mathbf{z}}_{t+h}$, which is not assigned any explicit target derived from the observation at $t{+}h$.

The Unrolled Action Head $H_{\bar\theta}$ shares the same architecture, except that 
the evolved state $\hat{\mathbf{z}}_{t+h}$ replaces the observation-derived 
conditioning tokens as the policy-conditioning input.
During training, it predicts the next-chunk velocity field:
\begin{equation}
    \hat{\mathbf{v}}_t^{(1)}
    =
    H_{\bar\theta}\bigl(\hat{\mathbf{z}}_{t+h},\mathbf{x}_{\tau_1}^{(1)},\tau_1\bigr).
    \label{eq:unrolled_head}
\end{equation}
The evolved state $\hat{\mathbf{z}}_{t+h}$ serves as its sole policy-conditioning 
input, while the noisy next-chunk action tokens and the flow-matching time $\tau_1$ 
are required by the action-generation objective.
The next-chunk loss propagates through the evolved state to the Online Action Head 
and the Context Adapter.

\subsection{Training Objective and Inference}
\label{sec:training}

Both action heads are trained with the flow matching 
objective~\cite{liu2022flow,lipman2022flow}.
For each chunk $j\in\{0,1\}$, we independently draw a flow-matching time $\tau_j$ 
and Gaussian noise $\boldsymbol{\epsilon}^{(j)}$, and construct the interpolated 
action tokens and the corresponding target velocity as
\begin{equation}
    \mathbf{x}_{\tau_j}^{(j)}
    =
    \tau_j\mathbf{A}_t^{(j)}
    +(1-\tau_j)\boldsymbol{\epsilon}^{(j)},
    \qquad
    \mathbf{v}_t^{(j)}
    =
    \mathbf{A}_t^{(j)}
    -\boldsymbol{\epsilon}^{(j)}.
    \label{eq:flow_path}
\end{equation}
Crucially, we sample $\tau_0$ and $\tau_1$ independently rather than constraining their relative noise levels. This decoupling exposes the unrolled head to $\hat{\mathbf{z}}_{t+h}$ across all noise combinations, encouraging the online head to produce a consistently informative evolved state regardless of the current denoising stage.
The training objective combines the velocity-prediction errors of the two chunks:
\begin{equation}
\begin{aligned}
    \mathcal{L}_{\mathrm{curr}}
    &=
    \mathbb{E}\left[
        \left\|
        \mathbf{v}_t^{(0)}
        -\hat{\mathbf{v}}_t^{(0)}
        \right\|_2^2
    \right],\\
    \mathcal{L}_{\mathrm{next}}
    &=
    \mathbb{E}\left[
        \left\|
        \mathbf{v}_t^{(1)}
        -\hat{\mathbf{v}}_t^{(1)}
        \right\|_2^2
    \right],\\
    \mathcal{L}_{\mathrm{total}}
    &=
    \mathcal{L}_{\mathrm{curr}}
    +\gamma\mathcal{L}_{\mathrm{next}},
\end{aligned}
\label{eq:objective}
\end{equation}
where $\gamma$ weights the contribution of next-chunk supervision.
Notably, the objective involves neither reconstruction of future observations nor 
alignment with future features.

The Unrolled Action Head shares the architecture of the Online Action Head but 
is excluded from gradient-based updates. 
Instead, its parameters are updated by an exponential moving average of the 
online head after each optimization step:
\begin{equation}
    \bar\theta
    \leftarrow
    m\bar\theta+(1-m)\theta,
    \label{eq:ema}
\end{equation}
where $m$ is the momentum coefficient. 
This design follows the practice of maintaining a slowly evolving copy of the 
model to provide reference predictions, as in target networks in reinforcement 
learning~\cite{mnih2015human,lillicrap2015continuous} and momentum encoders in 
self-supervised learning~\cite{he2020momentum,grill2020bootstrap,tarvainen2017mean}. 
The next-chunk loss does not directly update $\bar\theta$, but its gradient 
with respect to the evolved state is preserved and backpropagated to the online 
pathway. 
We ablate this design choice in Section~\ref{sec:experiments} 
(Table~\ref{tab:metaworld_ablation}).

At inference time, the policy uses the pretrained encoder, Context Adapter, and Online Action Head.
Starting from $\mathbf{x}_0^{(0)}\sim\mathcal{N}(\mathbf{0},\mathbf{I})$, we obtain the current action chunk by integrating the online head's predicted velocity field from $\tau_0=0$ to $\tau_0=1$.
Because the Unrolled Action Head is discarded after training, AdaPolicy introduces zero additional parameters or action-head evaluations at inference time.

\begin{figure*}[t]
\centering
\includegraphics[width=\textwidth, trim=1 49 100 1,clip]{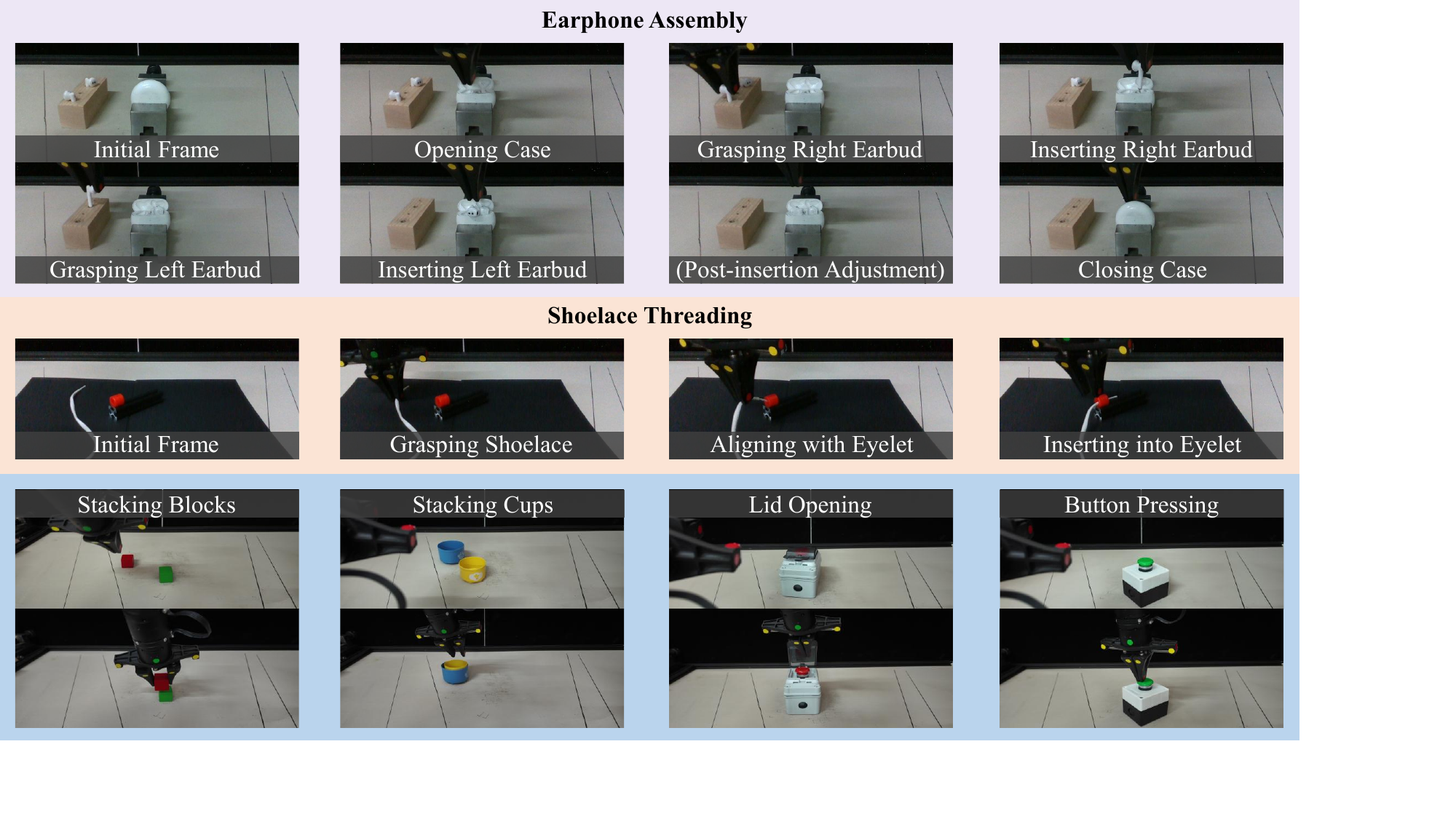} 
\caption{
\textbf{Visualization of real-world tasks.} 
Representative keyframes of the sub-tasks are shown for the two challenging 
millimeter-precision tasks (top: earphone assembly, shoelace threading); 
the remaining four tasks are shown below.
}
\label{fig:TaskVisAll}
\end{figure*}

\section{Experiments}
\label{sec:experiments}

We evaluate \textbf{AdaPolicy} to address four core questions:
(1) Does next-chunk supervision improve manipulation performance across simulation benchmarks and real-world tasks?
(2) Which components of the cascaded training scheme contribute to its effectiveness?
(3) Do the learned representations encode more information about future object states, and do they better support downstream 
planning?
(4) How does the next-chunk objective interact with the current-chunk objective during training?
We examine these questions through task-level evaluations, controlled ablations, future-position probes, and gradient analysis.

\begin{table}[h]
\centering
\caption{Success rates on real-world tasks. Each task is evaluated over 50 rollouts with randomized initial configurations.}
\label{tab:real_world_results}
\small
\begin{tabular}{lccccccc}
\toprule
Method 
& \begin{tabular}[c]{@{}c@{}}Earphone\\Assembly\end{tabular}
& \begin{tabular}[c]{@{}c@{}}Shoelace\\Threading\end{tabular}
& \begin{tabular}[c]{@{}c@{}}Stacking\\Blocks\end{tabular}
& \begin{tabular}[c]{@{}c@{}}Stacking\\Cups\end{tabular}
& \begin{tabular}[c]{@{}c@{}}Button\\Pressing\end{tabular}
& \begin{tabular}[c]{@{}c@{}}Lid\\Opening\end{tabular} 
& Avg. \\
\midrule
Baseline & 74 & 36 & 82 & 100 & 70 & 100 & 77.0 \\
AdaPolicy  & \textbf{86} & \textbf{50} & \textbf{94} & 100 & \textbf{88} & 100 & \textbf{86.3} \\
\bottomrule
\end{tabular}
\end{table}

\begin{table}[h]
\centering
\caption{Success rates on MetaWorld in comparison with prior methods, averaged over all 50 tasks.}
\label{tab:metaworld}
\small
\setlength{\tabcolsep}{6pt}
\begin{tabular}{lcccccc}
\toprule
Method & TinyVLA~\cite{tinyvla} & $\pi_0$~\cite{black2024pi_0} & SmolVLA~\cite{shukor2025smolvla} & VPP~\cite{hu2024video} & RoMAN-Flow~\cite{wang2026roman} & \textbf{AdaPolicy} \\
\midrule
Avg. & 31.6 & 50.8 & 68.2 & 68.2 & 72.8 & \textbf{81.2} \\
\bottomrule
\end{tabular}
\end{table}

\begin{table}[h]
\centering
\caption{
Average success rates on RoboTwin 2.0.
(a) Comparison on 50 tasks, evaluated under both clean and randomized 
scenes (50 rollouts per task).
(b) Ablation over backbones on 10 tasks, evaluated on clean scenes 
(50 rollouts per task).
}
\label{tab:robotwin}
\small
\begin{tabular}{lccccc}
\toprule
\multicolumn{6}{c}{(a) \emph{50 tasks, 50 clean + 500 randomized demos per task}} \\
\midrule
Method & $\pi_0$ & X-VLA & ABot-M0 & Baseline & AdaPolicy \\
\midrule
Avg. (\emph{clean}) & 65.9 & 72.8 & 81.2 & 64.4 & \textbf{84.92} \\
\midrule
Avg. (\emph{randomized}) & 58.4 & 72.8 & 80.4 & 64.5 & \textbf{84.04} \\
\bottomrule
\end{tabular}

\vspace{6pt}

\begin{tabular}{lccc}
\toprule
\multicolumn{4}{c}{(b) \emph{10 tasks, 50 clean + 200 randomized demos per task}} \\
\midrule
Backbone & Baseline & Baseline + future-feature sup. & AdaPolicy \\
\midrule
Avg. (DINOv3)      & 63.8 & 69.8 & \textbf{72.4} \\
Avg. (Qwen3-VL-2B) & 55.2 & 61.0 & \textbf{61.8} \\
\bottomrule
\end{tabular}
\end{table}

\subsection{Experimental Setup}
\label{sec:exp_setup}

\paragraph{Real-World Evaluation.}
We further evaluate AdaPolicy on six real-world manipulation tasks of varying 
difficulty, as illustrated in Figure~\ref{fig:TaskVisAll}.
The suite includes two high-precision tasks---earphone assembly and deformable 
shoelace threading---both requiring millimeter-level precision, together with four 
tabletop tasks of moderate difficulty.
We collect 1{,}000 demonstrations for earphone assembly and 400 for shoelace 
threading, while each of the remaining four tasks is provided with 80 demonstrations.
All tasks are evaluated over 50 trials.
To assess robustness to initial configurations, we randomize the poses of the 
earbuds and charging case in the assembly task, as well as the shoelace 
configuration and eyelet pose in the threading task.
Task protocols, demonstration collection, and evaluation procedures are detailed 
in the supplementary material.

\begin{table}[t]
\centering
\caption{
Success rates on the four LIBERO suites in comparison with prior methods.
}
\label{tab:libero}
\small
\begin{tabular}{lccccc}
\toprule
Method & Goal & Object & Spatial & Long & Avg. \\
\midrule
OpenVLA~\cite{openvla}         & 79.2 & 88.4 & 84.7 & 53.7 & 76.5 \\
SmolVLA~\cite{shukor2025smolvla} &	91.0 &	94.0	&93.0	&77.0&	88.8\\
DreamVLA~\cite{zhang2025dreamvla}       & 89.5 & 94.0 & 97.5 & 89.5 & 92.6 \\
GR00T-N1~\cite{gr00tn1}       & 94.4 & 97.6 & 93.0 & 90.6 & 93.9 \\
$\pi_0$~\cite{black2024pi_0}             & 96.8 & 98.8 & 95.8 & 85.2 & 94.1 \\
\midrule
Baseline              & 93.0 & 96.4 & 96.0 & 83.0 & 92.1 \\
\textbf{AdaPolicy} & 95.0 & 98.8 & 96.8 & 90.2 & \textbf{95.2} \\
\bottomrule
\end{tabular}
\end{table}

\paragraph{Simulation Benchmarks.}
We evaluate on MetaWorld~\cite{metaworld}, RoboTwin 2.0~\cite{robotwin},  and LIBERO~\cite{libero}.
On MetaWorld, we jointly train a single policy across 50 manipulation tasks, using 50 demonstration trajectories per task.
On RoboTwin~2.0, we consider two settings: the full 50-task setting, where a single policy is jointly trained with 50 clean and 500 randomized demonstrations per task, and a 10-task setting, where a policy is jointly trained with 50 clean and 200 randomized demonstrations per task.
For LIBERO, we evaluate on the Goal, Object, Spatial, and Long task suites.
Across all three benchmarks, we conduct 50 evaluation rollouts per task and report success rates.

\paragraph{Implementation Details.}
By default, the action chunk length is $h{=}32$ on RoboTwin 2.0 and $h{=}8$ on 
LIBERO and MetaWorld.
The next-chunk loss weight is $\gamma{=}0.8$ on LIBERO and MetaWorld, and 
$\gamma{=}0.5$ on RoboTwin 2.0.
The Unrolled Action Head is parameterized by an exponential 
moving average (EMA) of the Online Action Head, with decay 
$m{=}0.99$ on MetaWorld and LIBERO, and $m{=}0.98$ on RoboTwin~2.0.
We experiment with two pretrained encoders: the Qwen3-VL-2B-Instruct~\cite{bai2025qwen3} and DINOv3~\cite{simeoni2025dinov3}.
Both encoders are kept frozen during training, and DINOv3 is used as the default feature extractor unless otherwise specified.
All models are trained on NVIDIA RTX 5090 GPUs, and evaluation is 
conducted on NVIDIA RTX 4090 GPUs.
Because AdaPolicy operates purely on action tokens during the unroll without encoding future observations, training is highly efficient: jointly training the 50-task policy on RoboTwin~2.0 requires only about $60$ GPU hours.
At inference, actions are denoised using an ODE solver with 10 steps.

Our baseline shares the same architecture, training recipe, and 
evaluation protocol as AdaPolicy; the two differ only in whether 
cascaded next-chunk prediction is used, which isolates the effect 
of next-chunk supervision.
For comparison with explicit future supervision, the 
\emph{future-feature supervision} variant replaces next-chunk 
prediction with feature alignment: an additional decoder takes 
the evolved state as input and is trained with a cosine loss to 
match intermediate-layer features extracted from the future 
observation by a frozen DINOv3 encoder.
Configuration details are in Appendix~\ref{app:eval_setting}.

\subsection{Performance in Real World and Simulation}
\label{sec:main_results}

\paragraph{Real World.}
We evaluate on six real-world manipulation tasks of varying 
difficulty, ranging from tabletop rearrangement to 
millimeter-precision tasks.
As shown in Table~\ref{tab:real_world_results}, AdaPolicy raises 
the average success rate from 77.0\% to 86.3\%.
The improvements are concentrated on the more demanding tasks.
On Earphone Assembly, a multi-step task that requires 
millimeter-level precision over a long horizon, the success rate 
increases from 74\% to 86\%.
On Shoelace Threading, which validates manipulation of deformable 
objects, it increases from 36\% to 50\%.
Meanwhile, performance on the two saturated tasks (Stacking Cups 
and Lid Opening, both at 100\%) is maintained.
These results indicate that next-chunk supervision remains 
effective on challenging real-world tasks that demand 
high-precision manipulation.

\paragraph{Simulation.}
On MetaWorld, AdaPolicy achieves an average success rate of 81.2\% 
over all 50 tasks (Table~\ref{tab:metaworld}), outperforming 
compared methods, including RoMAN-Flow (72.8\%) and VPP (68.2\%).
On LIBERO, AdaPolicy improves over the baseline on all four suites, raising 
the average success rate from 92.1\% to 95.2\% 
(Table~\ref{tab:libero}); the largest gain appears on the Long 
suite ($+7.2$ points), which involves the longest horizons, 
consistent with the motivation that next-chunk supervision 
encourages the policy to anticipate upcoming states.
On RoboTwin~2.0, we jointly train a single model on all 50 tasks 
(Table~\ref{tab:robotwin}a).
AdaPolicy improves over the chunk-only baseline by 20.52 points; since the baseline shares exactly the same 
backbone, data, and training recipe, this gap is largely 
attributable to next-chunk supervision.
AdaPolicy also surpasses $\pi_0$ and X-VLA.
We further conduct an ablation on 10 tasks with two backbones, 
DINOv3 and Qwen3-VL-2B, to examine whether the benefit of 
next-chunk supervision depends on the choice of pretrained 
encoder (Table~\ref{tab:robotwin}b).
AdaPolicy outperforms both the baseline and explicit 
future-feature supervision under both backbones: with DINOv3, 
it improves over the baseline by 8.6 points and over 
future-feature supervision by 2.6 points; with Qwen3-VL-2B, 
the gains are 6.6 and 0.8 points, respectively.
These results suggest that next-chunk supervision yields reliable 
improvements across different task scales and adapts flexibly to both visual and vision-language backbones.

\subsection{Analysis of Key Design Choices}

\begin{figure*}[h]
\centering
\includegraphics[width=0.95\textwidth, trim=5 334 315 3,clip]{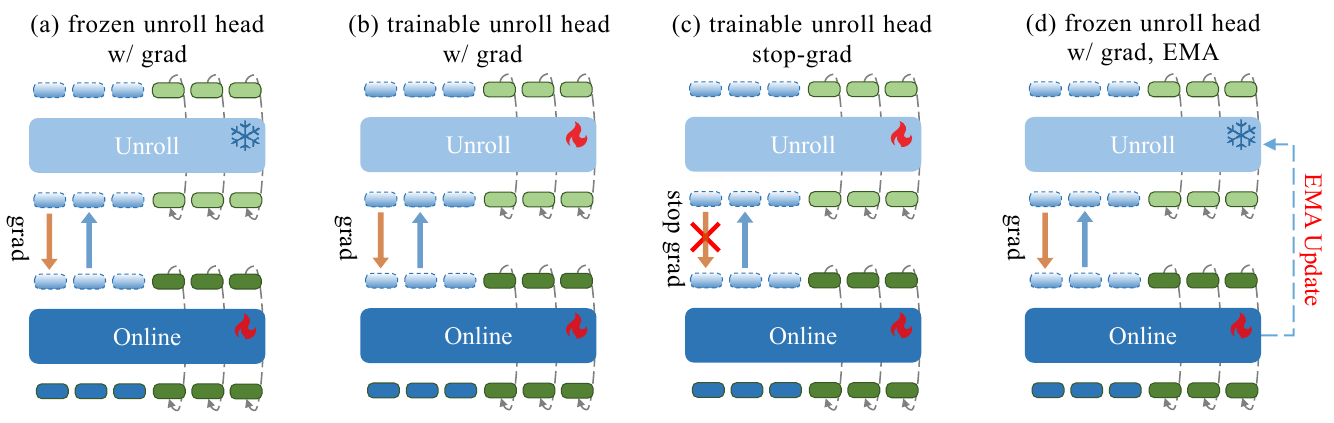} 
\caption{
\textbf{Illustration of the unrolled-head variants.}
Left: frozen unrolled head with gradients flowing to the online pathway. 
Middle: trainable unrolled head with gradients flowing to the online pathway. 
Right: trainable unrolled head with stop-gradient on the online 
pathway.
}
\label{fig:AblationCompare}
\end{figure*}

\begin{table}[h]
\centering
\caption{Ablation on MetaWorld (average success rate over 50 tasks). 
Next-chunk prediction surpasses both the extended-horizon baseline and 
explicit future-feature reconstruction; gradient flow through the evolved 
state and the EMA parameterization of the unrolled head are both critical.}
\label{tab:metaworld_ablation}
\small
\begin{tabular}{lcccc}
\toprule
Variant & $h$ & $\gamma$ & EMA $m$ & Avg. \\
\midrule
\multicolumn{5}{l}{\emph{w/o next-chunk prediction}} \\
\midrule
Baseline                              & 8  & --   & --    & 67.8 \\
Baseline (extended horizon)           & 16 & --   & --    & 74.3 \\
Baseline (extended horizon)           & 24 & --   & --    & 60.4 \\
Baseline + future-feature supervision & 8 & --   & --    & 68.8 \\
\midrule
\multicolumn{5}{l}{\emph{w/ next-chunk prediction}} \\
\midrule
\quad w/o condition$\to$action attention          & 8 & 0.8 & 0.99 & 74.9 \\
(a) \quad frozen unrolled head, w/o EMA                          & 8 & 0.8 & --   & 72.0 \\
(b) \quad trainable unrolled head          & 8 & 0.8 & --   & 69.4 \\
(c) \quad trainable unrolled head, stop-grad      & 8 & 0.8 & --  & 67.7 \\
(d) \quad AdaPolicy                        & 8 & 0.8 & 0.99 & \textbf{81.2} \\
\bottomrule
\end{tabular}
\end{table}

\paragraph{Design Analysis.}
Table~\ref{tab:metaworld_ablation} analyzes the key design choices of the 
cascaded objective on MetaWorld (average success over 50 tasks), and 
Figure~\ref{fig:AblationCompare} illustrates the corresponding 
gradient-flow configurations.
We highlight four findings.

First, the gains go beyond simply predicting a longer horizon.
Extending the chunk from $8$ to $16$ improves the baseline from 
$67.8$ to $74.3$, and further extending it to $24$ degrades 
performance to $60.4$---below the $h{=}8$ baseline---indicating that 
directly predicting a longer horizon quickly reaches its limit.
Our cascaded objective attains $81.2$ at the same supervision span 
and execution horizon as the $h{=}16$ baseline.
Direct $2h$-step prediction jointly denoises near-term and more 
uncertain far-future actions within a single pass.
In contrast, our cascaded design decouples the two flow-matching 
stages and routes subsequent-action supervision exclusively through 
the evolved state $\hat{\mathbf{z}}_{t+h}$.
This structure both insulates near-term velocity estimation from 
far-horizon noise and acts as an information bottleneck that shapes 
more predictive latent representations (consistent with the probing 
comparisons in Appendix~\ref{app:probe}).
Moreover, explicit future-feature supervision brings only a marginal 
gain over the baseline ($68.8$ vs.\ $67.8$).
Because our supervisory signal originates directly from the action 
stream, it naturally provides a control-aligned objective that 
focuses latent capacity on decision-relevant dynamics.

Second, the parameterization of the unrolled head matters. 
Reusing the online parameters without EMA drops the average success rate to 
$72.0$, and an independently trainable unrolled head further degrades it to 
$69.4$; both are clearly inferior to the slowly evolving EMA target 
($m{=}0.99$, $81.2$). 
This mirrors the classical role of target networks in stabilizing bootstrapped 
updates: the second-pass supervision is reliable only when its predictor 
$H_{\bar\theta}$ evolves more slowly than the representation it shapes.

Third, the improvement is primarily mediated by gradient flow 
through the evolved state, rather than by the auxiliary loss 
alone.
Applying stop-gradient to the evolved state $\hat{\mathbf{z}}_{t+h}$ 
yields performance nearly identical to the baseline 
($67.7$ vs.\ $67.8$), indicating that the auxiliary loss alone 
brings no measurable gain.
One possible explanation is that, without gradient flow, the 
next-chunk loss acts largely as an independent prediction objective 
that offers limited shaping of the control representation.
These results indicate that next-chunk supervision benefits the 
policy mainly by shaping the online pathway and the conditioning 
tokens $\mathbf{c}_t$. The auxiliary objective alone appears 
insufficient to account for the improvement.

Finally, the evolved state must actually inform action prediction.
Blocking the condition-to-action attention pathway in the second 
pass drops performance from $81.2$ to $74.9$.
This indicates that next-chunk supervision does not merely 
regularize the representation, but relies on the evolved state 
being accessible to downstream action tokens.

\paragraph{Hyperparameter Analysis.}
Table~\ref{tab:hyperparam} examines two key hyperparameters of 
the cascaded scheme: the next-chunk loss weight $\gamma$ and 
the EMA momentum $m$ of the Unrolled Action Head.
On MetaWorld, setting $\gamma{=}0$ recovers the baseline without 
next-chunk prediction, while all tested values from 
$0.5$ to $0.9$ bring clear gains over the baseline.
For the EMA momentum, performance improves as $m$ increases from 
$0.9$, peaking at $m{=}0.98$ on RoboTwin~2.0 (72.4\%) and 
remaining high at large $m$ on MetaWorld (81.4\% at 
$m{=}0.995$), indicating that maintaining a slowly evolving 
unrolled head benefits next-chunk supervision.

We conjecture that the EMA design helps for two reasons.
First, since the unrolled head is excluded from gradient updates, 
the next-chunk error can only be reduced through the evolved 
state, which encourages the online pathway to encode information 
useful for subsequent actions rather than directly co-adapting 
both heads.
Second, the unrolled head evolves more slowly than the online 
head, so the next-chunk target it defines remains stable over 
training instead of shifting at every optimization step, similar 
to the momentum teacher in self-supervised 
learning~\cite{he2020momentum,grill2020bootstrap,dino}.

\begin{table}[h]
\centering
\caption{
Hyperparameter sensitivity (average success rate).
(a) Loss weight $\gamma$ on MetaWorld with EMA $m{=}0.98$ ($^\dagger\gamma{=}0$ degenerates to the baseline without next-chunk prediction);
(b) EMA momentum $m$ on MetaWorld with $\gamma{=}0.8$;
(c) EMA momentum $m$ on RoboTwin~2.0.
}
\label{tab:hyperparam}
\small
\setlength{\tabcolsep}{5pt}
\begin{tabular}{lcccccc}
\toprule
\multicolumn{7}{l}{\textbf{(a)} Loss weight $\gamma$ on MetaWorld ($m{=}0.98$)} \\
\midrule
$\gamma$ & 0              & 0.5  & 0.6  & 0.7  & 0.8           & 0.9  \\
Avg.      & 67.8$^\dagger$ & 76.4 & 74.4 & 76.4 & \textbf{76.7} & 75.9 \\
\midrule
\multicolumn{7}{l}{\textbf{(b)} EMA momentum $m$ on MetaWorld ($\gamma{=}0.8$)} \\
\midrule
$m$       & 0.95 & 0.96 & 0.97 & 0.98 & 0.99 & 0.995          \\
Avg.      & 76.0 & 76.2 & 76.7 & 76.7 & 81.2 & \textbf{81.4}  \\
\midrule
\multicolumn{7}{l}{\textbf{(c)} EMA momentum $m$ on RoboTwin~2.0} \\
\midrule
$m$       & 0.9  & 0.95 & 0.98           & 0.99 & 0.995 &      \\
Avg.      & 59.6 & 68.2 & \textbf{72.4}  & 67.8 & 65.4  &      \\
\bottomrule
\end{tabular}
\end{table}

\paragraph{Cascade Unroll Depth Analysis.}

Our default design performs a single latent unroll, and a natural 
question is whether a deeper cascade further improves 
performance.
To this end, we extend the cascade with a second unroll that 
predicts the chunk $[t{+}2h,\,t{+}3h)$ conditioned on the 
twice-evolved latent state, using an EMA-updated head and an 
additional loss weight $\gamma_2$.
As shown in Table~\ref{tab:unroll_depth}, the conclusion is 
consistent across both benchmarks: two-step variants fall short 
of the one-step default.
On MetaWorld, the gap is $3.8$--$6.1$ points; on RoboTwin~2.0, 
adding a second unroll lowers the success rate by about $6$ 
points under both values of $\gamma$.
Interestingly, within the two-step variants, a smaller weight on 
the third-chunk loss tends to recover part of the drop.
On MetaWorld, reducing $\gamma_2$ from $0.8$ to $0.6$ raises the 
success rate from $75.1\%$ to $77.4\%$, although a very small 
$\gamma_2{=}0.4$ brings no further benefit ($75.6\%$); on 
RoboTwin~2.0, the variant with smaller weights ($\gamma{=}0.5$, 
$\gamma_2{=}0.25$) degrades less than the one with larger weights 
($\gamma{=}0.7$, $\gamma_2{=}0.3$).

This pattern suggests that the deeper supervision is noisier: 
a low weight limits its interference, while a high weight lets 
the noise offset the gains of next-chunk prediction.
On one hand, as the target moves further into the future, its 
dependence on the current observation weakens.
On the other hand, the twice-evolved state lies two prediction 
steps away from real observations, so the third-pass supervision 
is anchored at an off-distribution latent whose gradient 
interferes with the cleaner shaping signal from the first unroll.
A similar effect is known in model-based control with learned 
dynamics, where longer rollouts do not necessarily help as model 
errors compound over 
steps~\cite{mbpo,tdmpc}.
A single unroll thus balances anticipation depth against 
supervision reliability.

\begin{table}[h]
\centering
\caption{Effect of unroll depth on MetaWorld and RoboTwin 2.0 (average success 
rate).
$\gamma$ and $\gamma_2$ denote the loss weights of the first and second unroll 
steps.}
\label{tab:unroll_depth}
\small
\setlength{\tabcolsep}{3pt}
\begin{tabular}{lcccccccc}
\toprule
& \multicolumn{4}{c}{MetaWorld} & \multicolumn{4}{c}{RoboTwin 2.0} \\
\cmidrule(lr){2-5} \cmidrule(lr){6-9}
& 1-step & \multicolumn{3}{c}{2-step} & \multicolumn{2}{c}{1-step} & \multicolumn{2}{c}{2-step} \\
\cmidrule(lr){2-2} \cmidrule(lr){3-5} \cmidrule(lr){6-7} \cmidrule(lr){8-9}
& $\gamma{=}0.8$ & $\gamma{=}0.8$, $\gamma_2{=}0.4$ & $\gamma{=}0.8$, $\gamma_2{=}0.6$ & $\gamma{=}0.8$, $\gamma_2{=}0.8$ 
& $\gamma{=}0.5$ & $\gamma{=}0.7$ &$\gamma{=}0.5$, $\gamma_2{=}0.25$ & $\gamma{=}0.7$, $\gamma_2{=}0.3$ \\
\midrule
Avg. & \textbf{81.2} & 75.6 & 77.4 & 75.1 & \textbf{72.4} & 67.0 & 66.2 & 61.0 \\
\bottomrule
\end{tabular}
\end{table}

\paragraph{Flow-Matching Noise Schedule Analysis.}
Both the online and unrolled heads operate via flow matching, with their noise 
magnitudes governed by $1-\tau_0$ and $1-\tau_1$, respectively. A natural 
question is how the relative noise levels of the two chunks affect overall 
policy performance.
On both RoboTwin~2.0 and MetaWorld, we compare our default 
independent sampling ($\tau_0 \perp \tau_1$) against three constrained 
variants during training: (i) applying stronger noise to the next chunk than 
the current one ($1-\tau_1 > 1-\tau_0$), (ii) applying weaker noise to the 
next chunk ($1-\tau_1 < 1-\tau_0$), and (iii) sharing an identical noise 
timestep across both heads ($\tau_1 = \tau_0$). 
As shown in Table~\ref{tab:noise_schedule}, both benchmarks exhibit a 
consistent ranking across these strategies. 
Forcing the unrolled head to operate at a higher noise level than the online 
head severely limits the benefit of next-chunk supervision, yielding no 
observable gain over the baseline on RoboTwin~2.0, while coupling the two 
heads with a shared timestep ($\tau_1 = \tau_0$) also remains suboptimal. 
In contrast, keeping the next chunk less noisy than the current chunk 
consistently improves over the higher-noise and shared-timestep variants, 
whereas unconstrained independent sampling achieves the best overall 
performance on both benchmarks.

This asymmetry aligns with the signal-to-noise characteristics of cascaded denoising. When the next chunk is heavily corrupted ($1-\tau_1 > 1-\tau_0$), its velocity target is largely dominated by stochastic noise removal rather than fine-grained trajectory structure, which can introduce higher-variance gradients through $\hat{\mathbf{z}}_{t+h}$ that are less effective for guiding the online policy. Conversely, conditioning the unrolled head on cleaner next-chunk interpolations ($1-\tau_1 < 1-\tau_0$) offers a more deterministic, lower-variance supervisory signal that better supports policy learning, even when the current chunk is noisy. Finally, rather than restricting training to a coupled diagonal ($\tau_1 = \tau_0$) or half-plane, fully independent sampling outperforms constrained regimes by preserving the full marginal timestep distribution and exposing the evolved state to all combinations of $(\tau_0, \tau_1) \in [0, 1]^2$.

\begin{table}[h]
\centering
\caption{Effect of relative flow-matching noise levels between the unrolled head ($1-\tau_1$) and the online head ($1-\tau_0$) on RoboTwin 2.0 and MetaWorld.}
\label{tab:noise_schedule}
\begin{tabular}{lccc}
\toprule
Noise Sampling Strategy & Condition on $(\tau_0, \tau_1)$ & RoboTwin 2.0 & MetaWorld \\
\midrule
Baseline (w/o next-chunk) & -- & 63.8 & 67.8 \\
Higher next-chunk noise & $1-\tau_1 > 1-\tau_0$ ($\tau_1 < \tau_0$) & 63.8 & 76.2 \\
Lower next-chunk noise & $1-\tau_1 < 1-\tau_0$ ($\tau_1 > \tau_0$) & 69.2 & 79.0 \\
Shared noise timestep & $\tau_1 = \tau_0$ & -- & 75.4 \\
Independent sampling (AdaPolicy) & $\tau_0 \perp \tau_1$ (default) & \textbf{72.4} & \textbf{81.2} \\
\bottomrule
\end{tabular}
\end{table}

\paragraph{Next-Chunk Temporal Offset Analysis.}

By default, AdaPolicy predicts the action chunk immediately 
following the current one, i.e., with a temporal offset $d{=}h{=}32$.
To examine the sensitivity to this choice, we conduct experiments 
on RoboTwin~2.0 with the Qwen3-VL-2B backbone, training two 
additional variants with $d{=}16$ and $d{=}48$, which supervise 
the action chunks over $[t{+}16,\,t{+}48)$ and 
$[t{+}48,\,t{+}80)$, respectively, while keeping all other 
settings identical.
As reported in Table~\ref{tab:temporal_offset}, both variants 
still outperform the baseline (55.2\%), indicating that the gains 
of AdaPolicy do not hinge on a specific offset.
The default $d{=}32$ yields the best performance, which we 
attribute to a trade-off: a smaller offset overlaps with the current chunk and thus provides limited additional 
future information, whereas a larger offset makes the prediction 
target more uncertain and weakens the supervision signal.

\begin{table}[h]
\centering
\caption{Effect of the temporal offset $d$ of the next-chunk prediction on RoboTwin 2.0. The action chunk size is $h=32$, and the default setting is $d=32$.}
\label{tab:temporal_offset}
\begin{tabular}{ccc}
\toprule
Temporal Offset $d$ & Next-chunk Interval & Avg. \\
\midrule
16 & $[t+16,\ t+48)$ & 58.6 \\
32 (default) & $[t+32,\ t+64)$ & \textbf{61.8} \\
48 & $[t+48,\ t+80)$ & 59.4 \\
\bottomrule
\end{tabular}
\end{table}

\subsection{Representation Analysis}
\label{sec:representation}

We analyze the learned representations through complementary 
lenses: linear probing of future object motion, the geometry and 
attention patterns of the evolved state, and finally whether the 
representations support latent-space planning.

\paragraph{Probing Analysis.}
To examine whether next-chunk supervision shapes future-aware 
representations, we linearly probe the future positions of the 
target object from frozen features on two benchmarks: the 
\emph{stack-blocks-two} task of RoboTwin~2.0 and 10 MetaWorld 
tasks (\emph{door-close, sweep, stick-push, basketball, 
peg-insert-side, plate-slide-side, shelf-place, lever-pull, 
pick-out-of-hole, disassemble}, all v3).
To prevent partially noised action tokens from leaking ground-truth action information, we feed pure Gaussian noise as the action input when extracting the evolved state for probing.
Features are aggregated by global average pooling and fed 
into a linear head trained to predict the object position at 
several future horizons; details are in Appendix~\ref{app:probe}.
We probe two feature sources---the condition state and the 
evolved state---and compare AdaPolicy with the 
baseline and a variant trained with explicit future-feature 
supervision.
First, next-chunk supervision yields lower probe error.
Second, the evolved state is more decodable than the condition state, whereas the two are close for the baseline ($15.76$ vs.\ $16.38$ px), suggesting that it is the next-chunk gradient, rather than the architecture alone, that organizes future information into a linearly accessible format.
Third, explicit future-feature supervision improves over the 
baseline but remains behind AdaPolicy on both benchmarks 
($2.55$ vs.\ $2.48$ px on MetaWorld; $14.07$ vs.\ $9.45$ px on 
RoboTwin~2.0).

\begin{table}[h]
\centering
\caption{
Linear probing of future object positions (average test-set 
error in pixels over future horizons, $\downarrow$, mean$\pm$std 
over 3 seeds).
}
\label{tab:probe}
\small
\begin{tabular}{lclcc}
\toprule
Model & $h$ & Probed features & RoboTwin 2.0 & MetaWorld \\
\midrule
\multirow{2}{*}{Baseline} & \multirow{2}{*}{8}
 & condition state &16.38{\scriptsize$\pm$0.12} & 2.89{\scriptsize$\pm$0.09} \\
 &
 & evolved state   &15.76{\scriptsize$\pm$0.07} & 2.67{\scriptsize$\pm$0.05} \\
\midrule
\multirow{2}{*}{Baseline + future-feature sup.} & \multirow{2}{*}{8}
 & condition state &15.10{\scriptsize$\pm$0.19} & 2.78{\scriptsize$\pm$0.05} \\
 &
 & evolved state   &14.07{\scriptsize$\pm$0.23} & 2.55{\scriptsize$\pm$0.05} \\
\midrule
\multirow{2}{*}{Baseline + extended horizon} & \multirow{2}{*}{16}
 & condition state &15.28{\scriptsize$\pm$0.16} & 3.00{\scriptsize$\pm$0.07} \\
 &
 & evolved state   &14.06{\scriptsize$\pm$0.25} & 2.73{\scriptsize$\pm$0.09} \\
\midrule
\multirow{2}{*}{AdaPolicy} & \multirow{2}{*}{8}
 & condition state &10.20{\scriptsize$\pm$0.21} & 2.77{\scriptsize$\pm$0.07} \\
 &
 & evolved state   &\textbf{9.45}{\scriptsize$\pm$0.25} & \textbf{2.48}{\scriptsize$\pm$0.03} \\
\bottomrule
\end{tabular}
\end{table}

\paragraph{Evolved-state Alignment.}
We measure the cosine similarity between the evolved state $\hat{\mathbf{z}}_{t+h}$ and the condition states of the current ($\mathbf{c}_t$) and next ($\mathbf{c}_{t+h}$) chunks on MetaWorld (Table~\ref{tab:cosine}).
Without next-chunk supervision, the evolved state nearly collapses to an identity copy of the current condition state ($0.9977$), indicating minimal temporal progression across the online head.
With next-chunk supervision, while $\hat{\mathbf{z}}_{t+h}$ still retains strong anchor information from the current context ($0.9650$), its alignment with the future condition state $\mathbf{c}_{t+h}$ increases substantially ($0.9424$ vs.\ $0.6458$).
Notably, this directional shift toward the upcoming conditioning state emerges without any explicit future-observation targets, confirming that subsequent-action supervision implicitly encourages the latent representation to transition forward in time.
We additionally examine the attention mass allocated by condition tokens (queries) to action tokens (keys) inside the online head: averaged over DiT layers and denoising timesteps, it increases from $0.148$ to $0.195$ under next-chunk supervision.
Without a subsequent-action objective, the state pathway is not explicitly pressured to incorporate the ongoing action stream, which aligns with its near-identity preservation of the initial condition ($0.9977$).
Under next-chunk supervision, however, accurately forecasting subsequent actions depends on the trajectory currently being generated; this optimization pressure encourages condition tokens to aggregate more information from the evolving action tokens when forming $\hat{\mathbf{z}}_{t+h}$, consistent with an implicit, action-driven state update.

\begin{table}[h]
\centering
\caption{
Representation analysis on MetaWorld.
Left: cosine similarity between the evolved state 
$\hat{\mathbf{z}}_{t+h}$ and the condition states of the current 
and the next chunk.
Right: attention mass from condition tokens to action tokens 
inside the online head, averaged over DiT layers and 
flow-matching denoising timesteps.
}
\label{tab:cosine}
\setlength{\tabcolsep}{6pt}
\begin{tabular}{lccc}
\toprule
& \multicolumn{2}{c}{Cosine similarity} & Attention mass \\
\cmidrule(lr){2-3} \cmidrule(lr){4-4}
& $\cos(\hat{\mathbf{z}}_{t+h},\ \mathbf{c}_t)$ & $\cos(\hat{\mathbf{z}}_{t+h},\ \mathbf{c}_{t+h})$ & cond $\to$ action \\
& (current) & (next) & (avg.) \\
\midrule
Baseline  & 0.9977 & 0.6458 & 0.148 \\
AdaPolicy & 0.9650 & 0.9424 & 0.195 \\
\bottomrule
\end{tabular}
\end{table}

\paragraph{Latent-space Planning.}
To further test whether the learned representations capture action-dependent future information, we perform trajectory selection in the latent space following the CEM protocol of~\cite{maes2026leworldmodel}, with the planning block aligned to our action chunk size ($h{=}8$).
Without training an external predictor, we directly pass candidate $8$-step action chunks into the frozen Online Action Head at the clean endpoint ($\tau_0{=}1$) to obtain the evolved state $\hat{\mathbf{z}}_{t+8}$.
Candidates are ranked by the squared $\ell_2$ distance between their evolved state and that of a goal observation (extracted at $\tau_0{=}0$), iteratively updated over 30 CEM iterations, and replanned every 8 environment steps.
We evaluate on the $\text{gap}{=}16$ beacon setting of MetaWorld, where the target object starts $10.6$\,cm away on average, and report the rate of reaching within $2/3/5$\,cm of the target (Table~\ref{tab:cem_planning}).
AdaPolicy consistently improves over the baseline across all thresholds, roughly doubling the reach rate at the tighter criteria.
Because both policies share the same architecture and differ only in next-chunk supervision, this improvement suggests that the cascaded objective encourages the evolved state to better reflect future action consequences.
These results complement our policy evaluation and linear-probing analyses, offering additional evidence that next-chunk supervision helps shape future-aware representations.

\begin{table}[t]
\centering
\caption{Latent-space planning with CEM on MetaWorld.
We report the rate of reaching within $2/3/5$\,cm of the goal.
}
\label{tab:cem_planning}
\small
\begin{tabular}{lccc}
\toprule
Method & reach@2cm & reach@3cm & reach@5cm \\
\midrule
Baseline  & 6.0  & 8.2  & 17.8 \\
AdaPolicy & \textbf{12.8} & \textbf{15.5} & \textbf{29.5} \\
\bottomrule
\end{tabular}
\end{table}

\begin{figure}[h]
\centering
\includegraphics[width=0.75\textwidth, trim=1 1 1 25,clip]{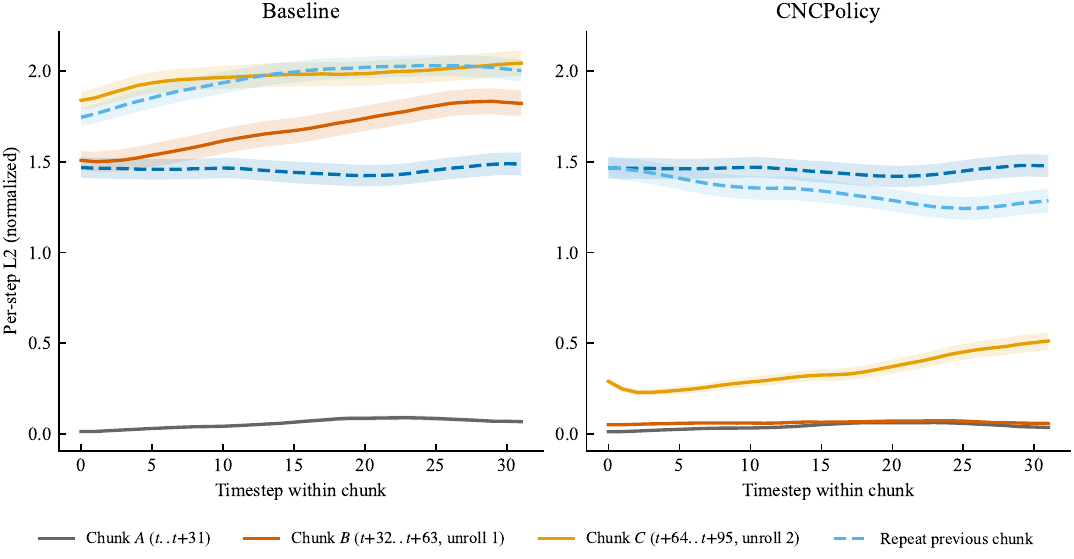} 
\caption{
\textbf{Per-step prediction error across successive chunks.}
Normalized L2 error on RoboTwin~2.0 for the current chunk 
($A$), the trained unroll ($B$), and the second unroll 
($C$), shown for the baseline (left) and AdaPolicy (right). 
Dashed lines mark the 
reference error of copying the previous ground-truth chunk.
}
\label{fig:fig_unroll_curves}
\end{figure}

\paragraph{Multi-step Unrolling.}
If the evolved state captures predictive structure about the future, it should remain informative when the cascade is rolled out further than during training.
AdaPolicy is trained with a single unroll, yet the cascaded structure can naturally be applied recursively: the evolved state produced by the second pass can condition a further unroll that predicts a subsequent chunk ($t{+}2h$ to $t{+}3h$, $C$) without additional training.
We examine this property on RoboTwin~2.0 by measuring the per-step normalized $\ell_2$ error between the predicted and ground-truth actions for each chunk.
As a reference, we also report the error of a naive persistence predictor that copies the ground-truth actions of the preceding chunk.
Figure~\ref{fig:fig_unroll_curves} shows the results.
For AdaPolicy, the prediction error stays below the persistence reference not only on the trained unroll ($B$) but also on the second unroll ($C$), where the error grows only mildly within the chunk.
In contrast, the baseline's error is comparable to the persistence reference on both chunks, suggesting that its evolved state carries little usable information about future actions.
This provides evidence that the learned anticipatory latent dynamics generalize beyond the unroll depth seen during training: the evolved state can be advanced one step further and still decode future actions better than the persistence reference.

\subsection{How Does Next-Chunk Supervision Affect Optimization?}

\begin{figure}[h]
\centering
\begin{subfigure}[b]{0.32\textwidth}
    \centering
    \includegraphics[width=\textwidth, trim=0 0 0 0, clip]{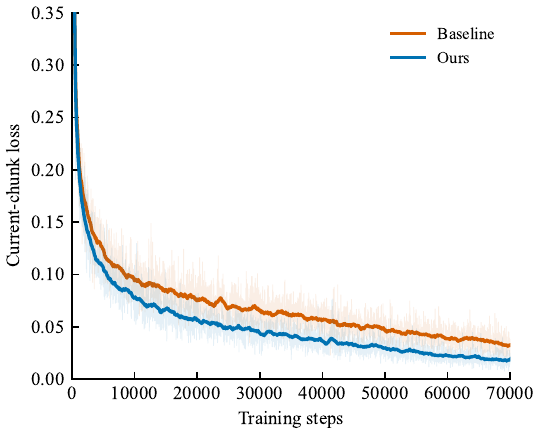}
    \caption{Effect of next-chunk prediction on training.
    Current-chunk loss with and without next-chunk supervision.
    }
    \label{fig:loss_curve_bj}
\end{subfigure}
\hfill
\begin{subfigure}[b]{0.32\textwidth}
    \centering
    \includegraphics[width=\textwidth, trim=1 1 1 1, clip]{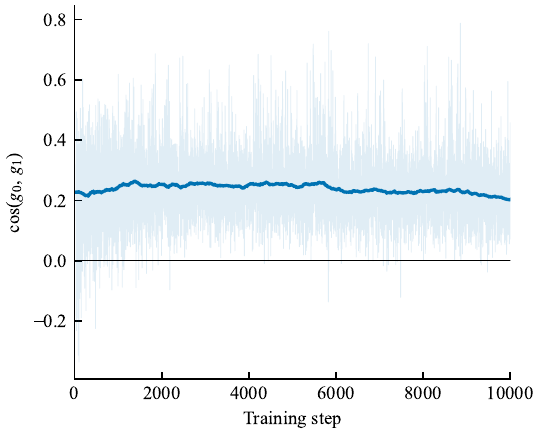}
    \caption{Cosine similarity between the current-chunk gradient $g_0$ and
    the next-chunk gradient $g_1$ during training.}
    \label{fig:GradientAnalysis}
\end{subfigure}
\hfill
\begin{subfigure}[b]{0.32\textwidth}
    \centering
    \includegraphics[width=\textwidth, trim=0 0 0 0, clip]{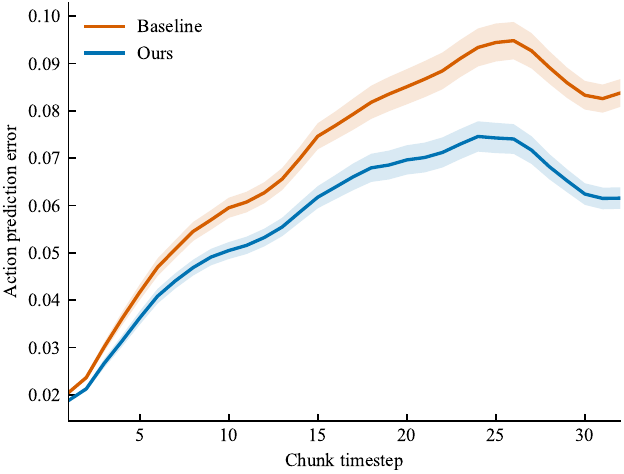}
    \caption{
    Per-timestep action prediction error within the current chunk 
    with and without next-chunk prediction.}
    \label{fig:chunk_error_robotwin}
\end{subfigure}
\caption{\textbf{How does next-chunk supervision affect optimization?}
(a) Next-chunk supervision consistently lowers the current-chunk training loss;
(b) the gradients of the two objectives stay positively correlated
(moving average around $0.2$), indicating no systematic gradient conflict;
(c) next-chunk prediction reduces the per-timestep action error throughout the chunk,
with the largest margin at the final step.}
\label{fig:optimization_analysis}
\end{figure}

\paragraph{Faster Convergence with Next-chunk Supervision.}

Figure~\ref{fig:loss_curve_bj} compares the current-chunk 
loss during training with and without next-chunk supervision on 
RoboTwin~2.0.
With next-chunk supervision, $\mathcal{L}_{\mathrm{curr}}$ 
decreases faster from early training, and the gap persists 
throughout the whole run.
This indicates that the next-chunk objective does not compete 
with current-chunk learning; rather, the gradient propagated 
through the evolved state serves as a useful shaping signal 
during the training process.
Combined with our observation-free latent unroll, this faster convergence enables AdaPolicy to achieve strong multi-task performance with high training efficiency.

\paragraph{Gradient Compatibility.}
An auxiliary objective tends to be more beneficial when its gradient does not 
systematically oppose that of the primary one~\cite{du2018adapting,yu2020gradient}. 
We therefore measure the cosine similarity between the current-chunk gradient 
$g_0=\nabla\mathcal{L}_{\mathrm{curr}}$ and the next-chunk gradient $g_1$, 
which is backpropagated through the evolved state $\hat{\mathbf{z}}_{t+h}$ onto 
the shared online pathway. 
As shown in Figure~\ref{fig:GradientAnalysis}, the similarity stays positive on average 
throughout training, with a moving average around $0.2$, suggesting that the 
next-chunk objective does not introduce systematic gradient conflict.

\paragraph{Action Prediction Error.}
To verify whether the anticipatory objective translates into more accurate action generation, we compare the current-chunk prediction error of models trained with and without next-chunk supervision at an identical number of training steps across all 50 tasks on RoboTwin~2.0. 
Both models are evaluated on $1{,}150$ randomly sampled test data points using a 10-step ODE solver to generate actions, under identical (observation, initial noise) pairs. 
We report the $\ell_2$ error between the generated and ground-truth actions at each of the $32$ chunk timesteps.

As shown in Figure~\ref{fig:chunk_error_robotwin}, AdaPolicy consistently achieves lower prediction error than the baseline across all $32$ timesteps, with clearly separated confidence bands after the initial few steps. 
While both models start with comparable error at the very beginning of the chunk, the baseline error accumulates rapidly as the horizon extends. 
In contrast, next-chunk supervision substantially suppresses this long-horizon drift, widening the performance advantage progressively toward the middle and later stages of the chunk. 
This behavior-level result directly complements our representation-level probing: the anticipatory objective not only organizes future-aware features, but also mitigates error accumulation over the chunk horizon to yield more accurate action trajectories.

\section{Conclusion}
\label{sec:conclusion}

This work questions a common premise behind future-aware policy learning: that anticipation is typically taught through targets derived from future observations.
We showed that the action signal itself can be harnessed as a source of supervision for building anticipation into policies.
AdaPolicy realizes this idea via cascaded next-chunk prediction, converting subsequent-action errors into an implicit shaping signal for future-aware representations.
The framework is architecturally simple, requiring neither future-observation 
reconstruction nor feature alignment.
Because the latent unroll operates strictly during training, AdaPolicy remains plug-and-play and incurs zero inference-time overhead compared to a standard chunk-based policy.
Extensive evaluations across simulation benchmarks and millimeter-precision real-world tasks, supported by representation probing and latent planning analyses, demonstrate consistent and substantial gains.
More broadly, our findings suggest that the action stream itself offers a viable signal for shaping anticipatory representations as a natural byproduct of control-oriented learning.

\bibliographystyle{unsrtnat}   
\bibliography{references}

\newpage
\appendix

\section{Comparison with explicit future-supervision methods}

\begin{figure}[h]
\centering
\includegraphics[width=\textwidth, trim=5 345 80 1,clip]{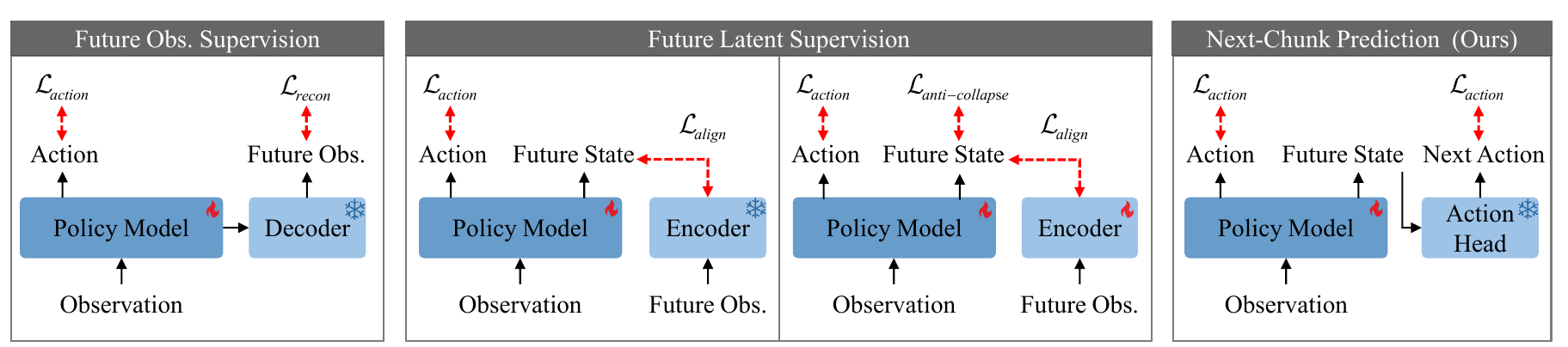} 
\caption{
\textbf{Comparison of future-supervision architectures.}
Existing methods supervise a predicted future state with 
explicit targets derived from future observations, either by 
pixel reconstruction or by feature alignment.
AdaPolicy instead derives supervision from the next action 
chunk, requiring no future-observation targets.
}
\label{fig:CompareOverview}
\end{figure}

Figure~\ref{fig:CompareOverview} contrasts AdaPolicy with the two 
main families of explicit future supervision.
\emph{Future-observation supervision} (left) trains a decoder on 
top of the policy model to reconstruct the future observation, 
using pixel reconstruction as the target; such targets devote 
substantial capacity to appearance details that are peripheral to 
control.
\emph{Future-latent supervision} (middle) instead aligns the 
predicted future state with features of the future observation, 
and comes in two forms.
The first freezes the feature extractor to keep the target fixed, 
which introduces a separate encoder and prevents the target 
representation from adapting to the policy.
The second trains the feature extractor jointly, which removes 
this restriction but exposes the pipeline to representation 
collapse, requiring dedicated anti-collapse losses or multi-stage 
training recipes to remain stable.
In contrast, AdaPolicy (right) derives the supervision signal 
from the next action chunk rather than future observations: the 
evolved state is shaped by a next-chunk action loss applied 
through an EMA-updated action head, requiring neither an explicit 
future target, an auxiliary alignment loss, nor a multi-stage 
training schedule.

\begin{table}[h]
\centering
\caption{
Training cost (batch size 128) under two 
hidden sizes.
AdaPolicy adds only modest computation over the baseline, whereas 
explicit future-feature supervision nearly doubles the per-sample 
FLOPs.
}
\label{tab:compute}
\small
\setlength{\tabcolsep}{6pt}
\begin{tabular}{lcccc}
\toprule
Variant & ms/step & Memory & Params & GFLOPs \\
\midrule
\multicolumn{5}{l}{\textbf{(a)} Hidden size $512$} \\
\midrule
AdaPolicy   & 1238  & 17.1\,GB & 93.1M  & 648.2 \\
Chunk-32 baseline        & 1236  & 13.7\,GB & 93.1M  & 611.5 \\
Chunk-64 baseline        & 1263  & 14.9\,GB & 93.1M  & 623.6 \\
Future-feature sup.      & 2415 & 17.2\,GB & 111.7M & 1163.8 \\
\midrule
\multicolumn{5}{l}{\textbf{(b)} Hidden size $768$} \\
\midrule
AdaPolicy   & 1401 & 23.9\,GB & 208.0M & 807.4 \\
Chunk-32 baseline        & 1350 & 18.7\,GB & 208.0M & 724.8 \\
Chunk-64 baseline        & 1381 & 20.5\,GB & 208.0M & 752.0 \\
Future-feature sup.      & 2561 & 23.0\,GB & 248.5M & 1316.7 \\
\bottomrule
\end{tabular}
\end{table}

\paragraph{Training cost.}
Table~\ref{tab:compute} reports the per-step training cost of 
each variant under two hidden sizes.
AdaPolicy increases the per-sample FLOPs over the chunk-32 
baseline by only $6\%$ (hidden $512$) and $11\%$ (hidden $768$), 
with identical parameter count; the overhead comes from the 
additional forward pass of the EMA-updated unrolled head, which 
is removed entirely at inference.
By contrast, explicit future-feature supervision requires an 
extra decoder and nearly doubles the per-sample FLOPs 
($1163.8$ vs.\ $648.2$ at hidden $512$), while also increasing 
the parameter count by $20\%$.

\section{Real-World Experiment Details}

\paragraph{Implementation details.}
Real-world experiments are conducted on an AgileX PiPer robotic arm, a 
6-DoF manipulator equipped with a parallel gripper. The action space 
consists of absolute joint positions, and the gripper is controlled with 
continuous values.
Visual observations are provided by two Intel RealSense D435 cameras 
streaming at 30\,Hz, and images are resized to $256 \times 256$ before 
being fed to the policy.
We use DINOv3 as the visual backbone.
The policy predicts action chunks of size 32, and the robot operates at 
a control frequency of 30\,Hz.
The weight $\gamma$ of the next-chunk prediction loss is set to $0.8$, 
and the momentum coefficient $m$ is set to $0.98$.
All real world experiments use identical data preprocessing and 
postprocessing pipelines.
This controlled setting helps isolate the effect of next-chunk prediction 
from differences in data handling.

\paragraph{Asynchronous inference.}
To prevent execution pauses caused by inference latency, we adopt an 
asynchronous inference scheme that decouples model inference from action 
execution via a multi-threaded design:
\begin{itemize}
    \item \textbf{Inference thread.} Runs the model to produce a chunk of 
    32 predicted actions, of which the first 16 are pushed into the 
    execution queue.
    \item \textbf{Execution thread.} Continuously dequeues actions from 
    the execution queue and dispatches them to the robotic arm at the 
    control frequency of 30~Hz.
\end{itemize}

The two threads communicate through a shared action queue with the 
following coordination protocol.
When the number of remaining actions in the queue drops to 8, the 
inference thread is triggered to compute the new action chunk in parallel 
with ongoing execution.
Once the new chunk becomes available, the execution queue is cleared and 
refilled with the latest 16 predicted actions, so that the robot always 
acts on the most up-to-date prediction.
This design hides inference latency behind action execution and ensures 
the continuity of motion.

\begin{figure}[h]
\centering
\includegraphics[width=0.6\textwidth, trim=5 345 597 5,clip]{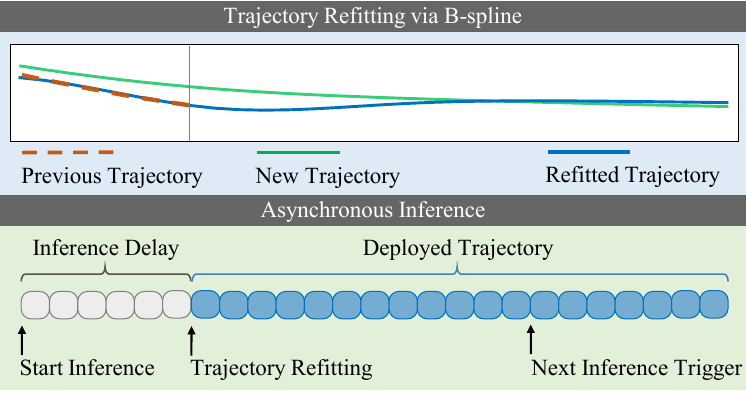} 
\caption{
\textbf{Real-world deployment with asynchronous inference.} 
Inference of the action chunk overlaps with the execution of the current 
one, and B-spline trajectory refitting smooths the transition across 
chunk boundaries.
}
\label{fig:BsplineAsync}
\end{figure}

\paragraph{Trajectory refitting across chunk boundaries.}
Naively switching from the remaining actions of the previous chunk to a 
newly predicted one can introduce discontinuities at the chunk boundary.
To smooth this transition, we refit each new action chunk with a B-spline~\cite{abpolicy} 
before execution (Figure~\ref{fig:BsplineAsync}):
(1)~the 16 predicted actions are first fitted with a cubic B-spline with 
8 control points;
(2)~the first 3 control points are then adjusted so that the resulting 
curve agrees with the already-executed portion of the previous chunk's 
trajectory, which mitigates abrupt jumps at the chunk boundary;
(3)~the trajectory is reconstructed from the updated control points and 
dispatched to the execution thread.
Since B-spline fitting only involves solving a small least-squares 
problem over a handful of control points, the refitting adds negligible 
computation and introduces no additional latency.

\begin{figure}[h]
\centering
\includegraphics[width=\textwidth, trim=5 230 5 1,clip]{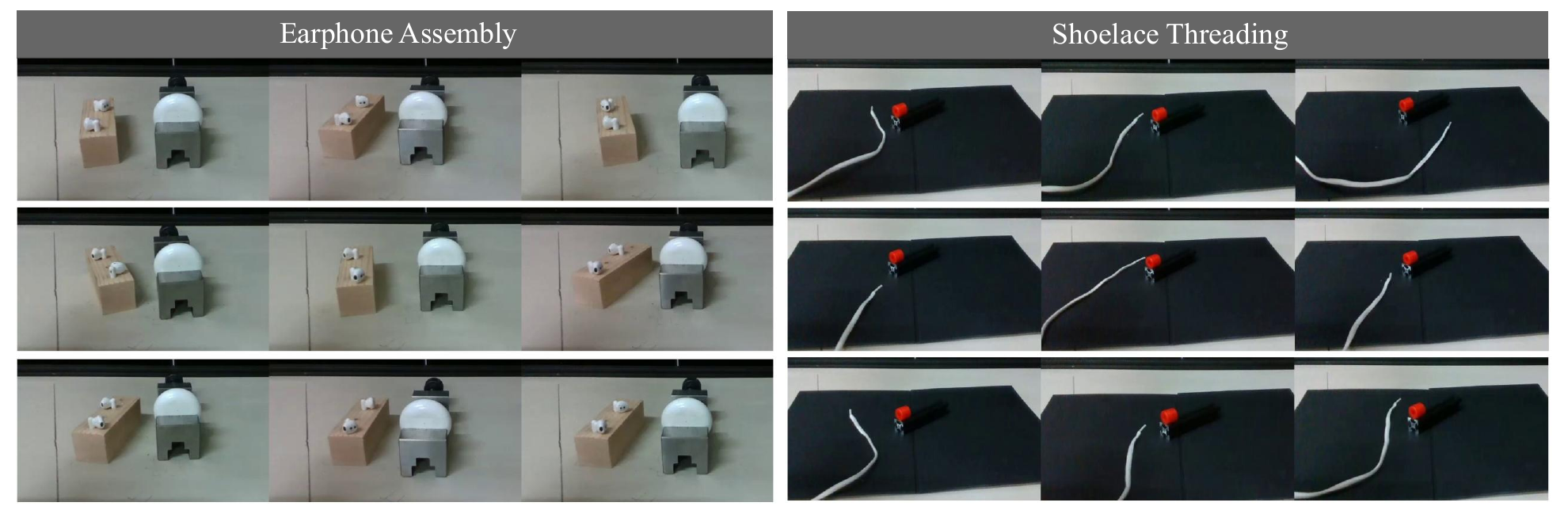} 
\caption{
Examples of randomized initial configurations in real-world 
evaluation.
}
\label{fig:RandomInitPic}
\end{figure}

\begin{figure}[h]
\centering
\includegraphics[width=0.6\columnwidth, trim=10 80 10 40,clip]{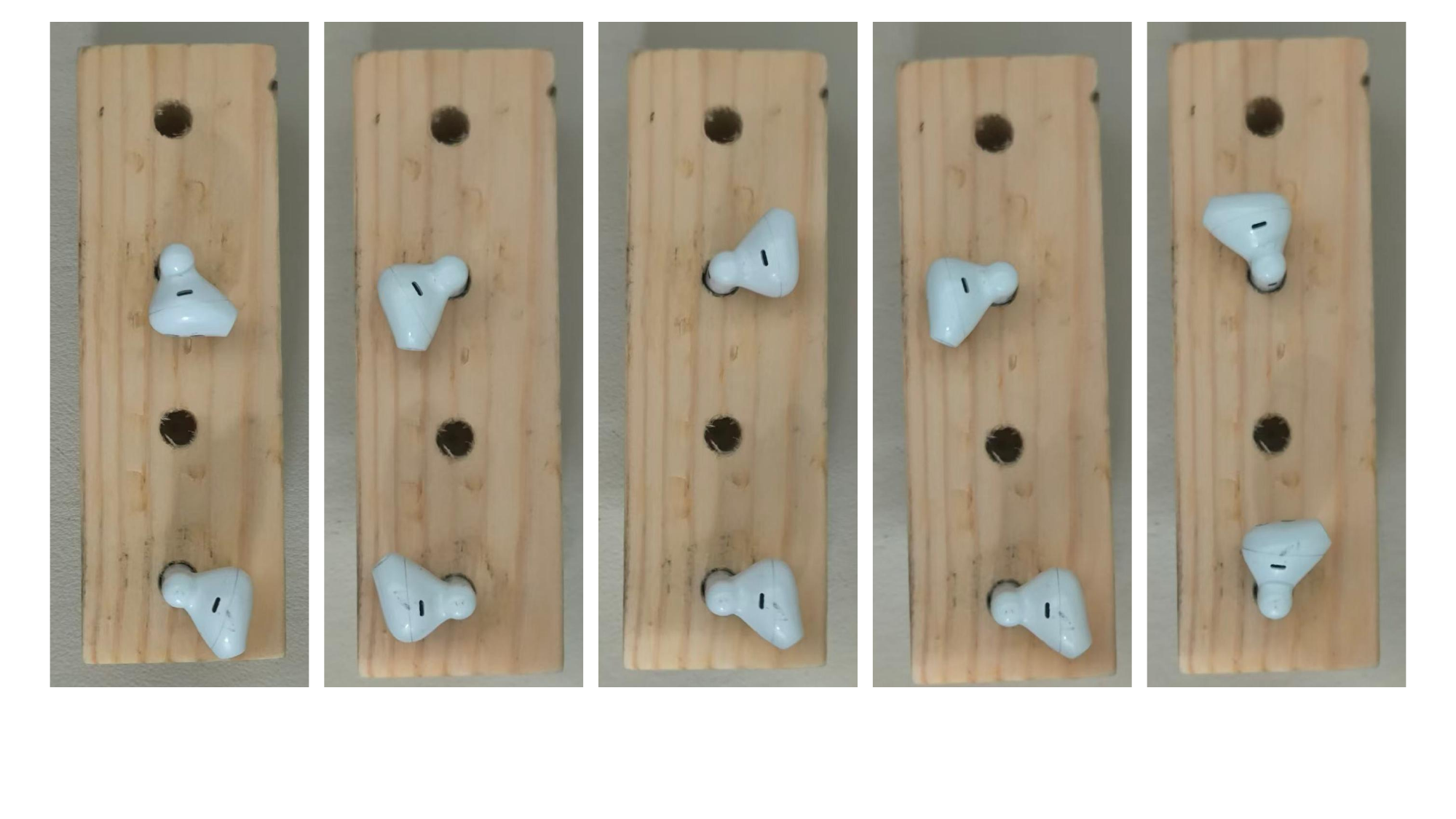}
\caption{Visualization of the random initial configurations in the earphone assembly task.}
\label{fig:InitialState_of_erji}
\end{figure}

\paragraph{Experimental Setup}

As illustrated in Figure~\ref{fig:RandomInitPic} and Figure~\ref{fig:InitialState_of_erji}, we present several examples of initialization randomness. 
During our real-world evaluations, the placements of both the wooden support block holding the earphones and the charging case are fully randomized. 
Specifically, the position of the charging case varies within an approximate $5 \times 5$~cm area. 
To further illustrate the precision required by this task, we measured the relevant components with a vernier caliper: the diameter of the earbud is 6.5~mm, while that of the corresponding charging slot is 7.4~mm. 
Such a tight tolerance demands millimeter-level precision throughout the insertion process. 
Moreover, due to the random initial orientation of each earbud, the gripper approaches and grasps it from varying poses, which inevitably introduces variability in the post-grasp configuration. The policy is therefore required to adapt to this state-dependent randomness and dynamically adjust its insertion trajectory accordingly.

\begin{figure}[h]
\centering
\includegraphics[width=\textwidth, trim=5 180 5 1,clip]{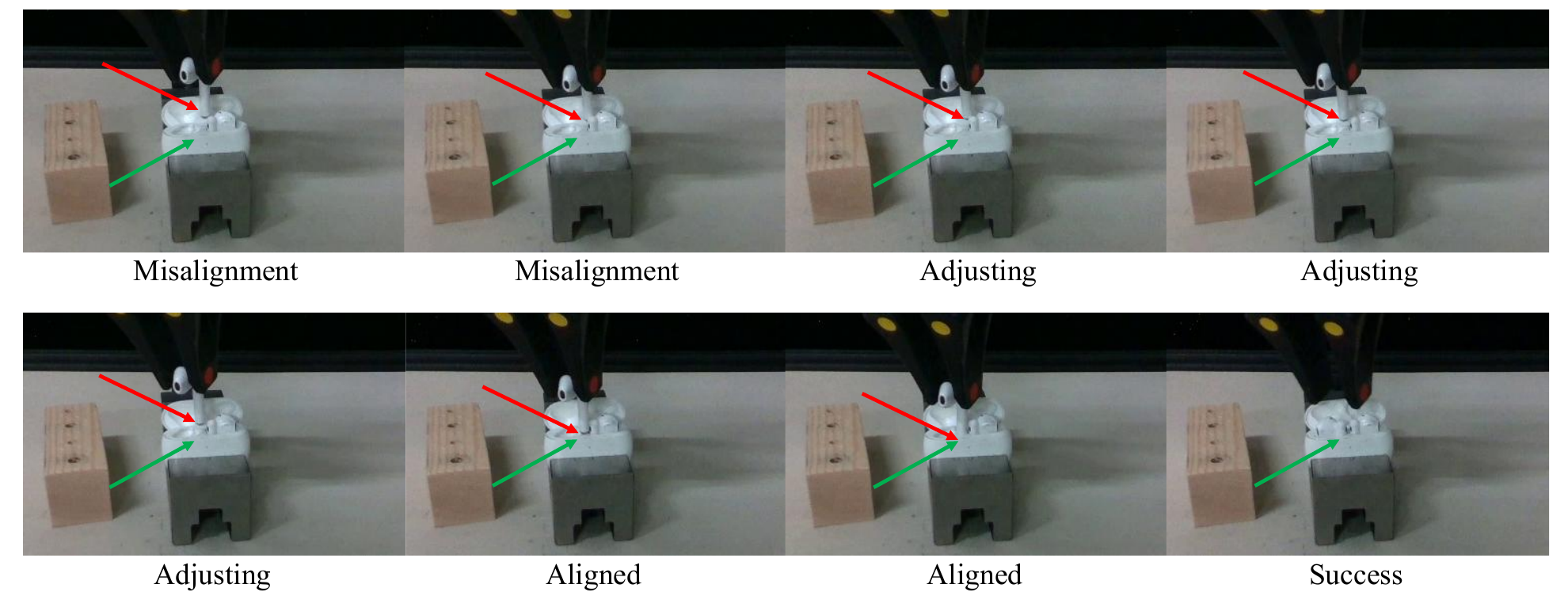} 
\caption{\textbf{Recovery from sub-optimal states on the earphone assembly task.} 
Red arrows indicate the earbud tip, and green arrows mark the insertion slot 
of the charging case. 
Starting from a misaligned insertion attempt, the policy progressively 
re-aligns the earbud and successfully inserts it into the case.}
\label{fig:RecoveryErji}
\end{figure}

\begin{figure}[h]
\centering
\includegraphics[width=\textwidth, trim=5 204 5 1,clip]{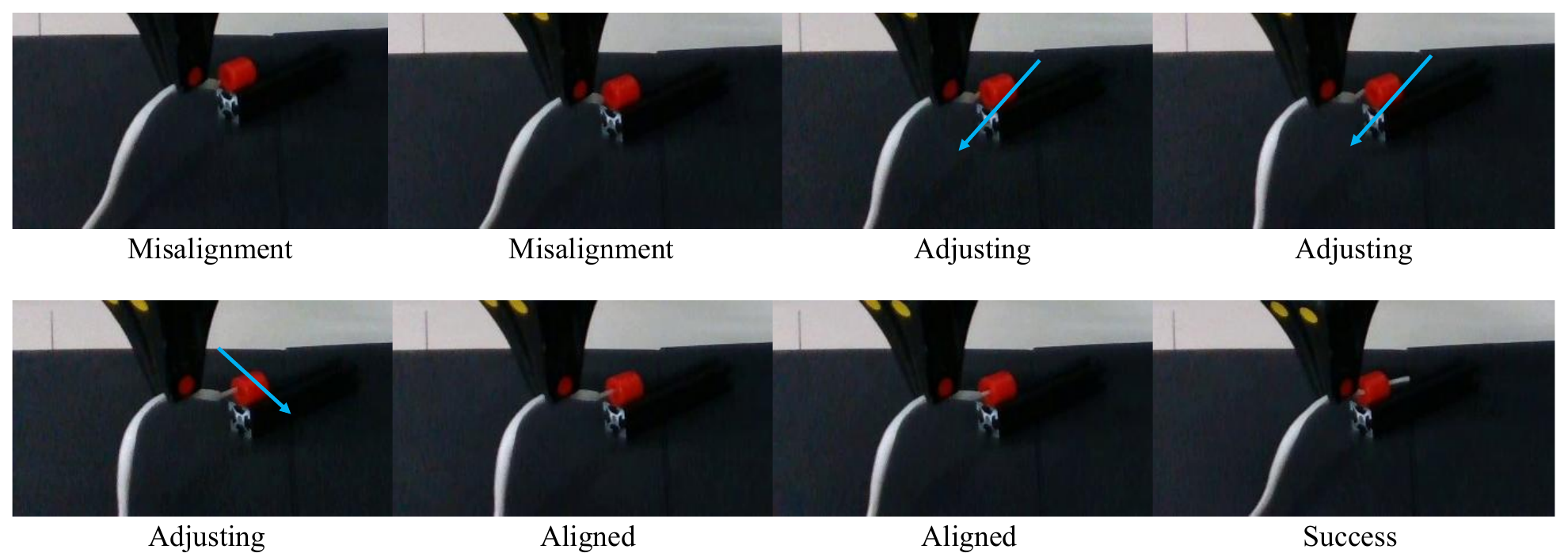} 
\caption{\textbf{Recovery from sub-optimal states on the shoelace threading task.} 
Blue arrows denote the adjustment direction. 
Starting from a misaligned initial attempt, the policy progressively re-aligns the shoelace and successfully threads it through the eyelet.}
\label{fig:RecoveryShoe}
\end{figure}

\paragraph{Recovery from sub-optimal states.}
On the earphone assembly and shoelace threading tasks, we additionally 
include human-in-the-loop corrections~\cite{hilserl} in the training data 
(the other real-robot tasks use teleoperated demonstrations only). 
Correction data amounts to 20\% of each task's dataset: half is collected 
by rolling out the Baseline policy and half by rolling out AdaPolicy. 
During collection, a human operator takes over whenever the policy enters 
a failure-prone configuration (e.g., a misaligned insertion attempt) and 
demonstrates the recovery behavior; the resulting takeover segments are 
added to the teleoperation dataset and used in joint training.
To ensure a fair comparison, both the Baseline and AdaPolicy are 
trained on the same combined dataset, guaranteeing identical data 
distributions across methods.
Figures~\ref{fig:RecoveryErji} and~\ref{fig:RecoveryShoe} show two 
representative recovery behaviors of AdaPolicy on the earphone assembly 
and shoelace threading tasks. Starting from a misaligned attempt, the 
policy progressively re-aligns the object and completes the insertion or 
threading, instead of repeating the failed motion or drifting away from 
the target.

\section{Probing Details}
\label{app:probe}

\paragraph{Setup and protocol.}

For each feature source (the condition state and the evolved 
state), token features are aggregated by global average pooling 
and fed into a linear probe consisting of a LayerNorm followed by 
a single linear layer, trained for 50 epochs with an $\ell_2$ 
loss while all policy parameters remain frozen.
Episodes are split at the episode level with a 3:2 ratio (on 
RoboTwin~2.0, 30 training and 20 test episodes for the 
\emph{stack-blocks-two} task), and stationary segments in which 
the target object does not move are removed, preventing the probe 
from solving the task by copying the current position.
The probe predicts the object position at horizons $t{+}16$, 
$t{+}32$, and $t{+}48$ on RoboTwin~2.0, and at $t{+}4$, $t{+}8$, 
$t{+}12$, and $t{+}16$ on MetaWorld; we report the test-set error 
between predicted and ground-truth positions in pixels, averaged 
over 3 seeds.
The \emph{stack-blocks-two} task is chosen because the target 
object undergoes large displacements, making future-position 
prediction non-trivial, while the 10 MetaWorld tasks are selected 
to cover diverse object interaction types.

\begin{table}[h]
\centering
\caption{
Linear probing of future object displacement on the RoboTwin 2.0 
\emph{stack-blocks-two} task (test-set midpoint error in pixels, $\downarrow$, 
mean$\pm$std over 3 seeds).
}
\label{tab:probe}
\small
\begin{tabular}{llcccc}
\toprule
Model & Probed features & $t{+}16$ & $t{+}32$ & $t{+}48$ & Avg. \\
\midrule
\multirow{2}{*}{Baseline ($h=32$)}
 & condition state & 11.85{\scriptsize$\pm$0.13} & 19.00{\scriptsize$\pm$0.41} & 18.29{\scriptsize$\pm$0.22} & 16.38{\scriptsize$\pm$0.12} \\
 & evolved state & 11.60{\scriptsize$\pm$0.13} & 18.05{\scriptsize$\pm$0.05} & 17.62{\scriptsize$\pm$0.27} & 15.76{\scriptsize$\pm$0.07} \\
\midrule
\multirow{2}{*}{Baseline ($h=32$) + future-feature sup.}
 & condition state & 10.95{\scriptsize$\pm$0.18} & 16.93{\scriptsize$\pm$0.59} & 17.43{\scriptsize$\pm$0.17} & 15.10{\scriptsize$\pm$0.19} \\
 & evolved state & 10.36{\scriptsize$\pm$0.10} & 15.75{\scriptsize$\pm$0.70} & 16.11{\scriptsize$\pm$0.24} & 14.07{\scriptsize$\pm$0.23} \\
\midrule
\multirow{2}{*}{Baseline ($h=64$)}
 & condition state & 11.02{\scriptsize$\pm$0.17} & 17.26{\scriptsize$\pm$0.67} & 17.57{\scriptsize$\pm$0.08} & 15.28{\scriptsize$\pm$0.16} \\
 & evolved state & 10.41{\scriptsize$\pm$0.09} & 16.06{\scriptsize$\pm$0.82} & 15.70{\scriptsize$\pm$0.02} & 14.06{\scriptsize$\pm$0.25} \\
\midrule
\multirow{2}{*}{AdaPolicy}
 & condition state & 8.34{\scriptsize$\pm$0.62} & 12.25{\scriptsize$\pm$0.85} & 10.00{\scriptsize$\pm$0.41} & 10.20{\scriptsize$\pm$0.21} \\
 & evolved state & \textbf{8.05}{\scriptsize$\pm$0.60} & \textbf{10.87}{\scriptsize$\pm$0.62} & \textbf{9.44}{\scriptsize$\pm$0.37} & \textbf{9.45}{\scriptsize$\pm$0.25} \\
\bottomrule
\end{tabular}
\end{table}

\begin{table}[h]
\centering
\caption{
Linear probing of future object displacement on 10 MetaWorld tasks 
(test-set midpoint error in pixels, $\downarrow$, mean$\pm$std over 
3 seeds). Protocol follows Table~\ref{tab:probe}.
}
\label{tab:probe_ablation}
\small
\begin{tabular}{llccccc}
\toprule
Model & Probed features & $t{+}$4 & $t{+}$8 & $t{+}$12 & $t{+}$16 & Avg. \\
\midrule
\multirow{2}{*}{Baseline ($h=8$)}
 & condition state & 1.44{\scriptsize$\pm$0.03} & 2.54{\scriptsize$\pm$0.06} & 3.42{\scriptsize$\pm$0.10} & 4.16{\scriptsize$\pm$0.15} & 2.89{\scriptsize$\pm$0.09} \\
 & evolved state & \textbf{1.33}{\scriptsize$\pm$0.02} & 2.32{\scriptsize$\pm$0.04} & 3.16{\scriptsize$\pm$0.06} & 3.87{\scriptsize$\pm$0.09} & 2.67{\scriptsize$\pm$0.05} \\
\midrule
\multirow{2}{*}{Baseline ($h=8$) + future-feature sup.}
 & condition state & \textbf{1.42}{\scriptsize$\pm$0.02} & \textbf{2.48}{\scriptsize$\pm$0.05} & 3.29{\scriptsize$\pm$0.06} & 3.93{\scriptsize$\pm$0.06} & 2.78{\scriptsize$\pm$0.05} \\
 & evolved state & \textbf{1.33}{\scriptsize$\pm$0.04} & \textbf{2.26}{\scriptsize$\pm$0.06} & 2.99{\scriptsize$\pm$0.06} & 3.61{\scriptsize$\pm$0.05} & 2.55{\scriptsize$\pm$0.05} \\
\midrule
\multirow{2}{*}{Baseline ($h=16$)}
 & condition state & 1.60{\scriptsize$\pm$0.04} & 2.72{\scriptsize$\pm$0.06} & 3.55{\scriptsize$\pm$0.10} & 4.12{\scriptsize$\pm$0.11} & 3.00{\scriptsize$\pm$0.07} \\
 & evolved state & 1.48{\scriptsize$\pm$0.05} & 2.48{\scriptsize$\pm$0.07} & 3.22{\scriptsize$\pm$0.09} & 3.73{\scriptsize$\pm$0.15} & 2.73{\scriptsize$\pm$0.09} \\
\midrule
\multirow{2}{*}{AdaPolicy (trainable unrolled head)}
 & condition state & 1.50{\scriptsize$\pm$0.03} & 2.58{\scriptsize$\pm$0.04} & 3.43{\scriptsize$\pm$0.05} & 4.04{\scriptsize$\pm$0.08} & 2.89{\scriptsize$\pm$0.05} \\
 & evolved state & 1.42{\scriptsize$\pm$0.04} & 2.36{\scriptsize$\pm$0.05} & 3.04{\scriptsize$\pm$0.05} & 3.62{\scriptsize$\pm$0.06} & 2.61{\scriptsize$\pm$0.05} \\
 \midrule
 \multirow{2}{*}{AdaPolicy}
 & condition state & 1.51{\scriptsize$\pm$0.03} & 2.56{\scriptsize$\pm$0.06} & \textbf{3.26}{\scriptsize$\pm$0.09} & \textbf{3.75}{\scriptsize$\pm$0.11} & \textbf{2.77}{\scriptsize$\pm$0.07} \\
 & evolved state & 1.43{\scriptsize$\pm$0.02} & 2.35{\scriptsize$\pm$0.03} & \textbf{2.88}{\scriptsize$\pm$0.03} & \textbf{3.24}{\scriptsize$\pm$0.04} & \textbf{2.48}{\scriptsize$\pm$0.03} \\
\bottomrule
\end{tabular}
\end{table}

\paragraph{Probing analysis.}
We probe on two benchmarks: the \emph{stack-blocks-two} task of 
RoboTwin~2.0 and 10 MetaWorld tasks (\emph{door-close, sweep, 
stick-push, basketball, peg-insert-side, plate-slide-side, 
shelf-place, lever-pull, pick-out-of-hole, disassemble}, all v3).
We probe two feature sources---the condition state and the evolved 
state---and compare AdaPolicy with the chunk-only baseline and a 
variant trained with explicit future-feature supervision.

Tables~\ref{tab:probe} and~\ref{tab:probe_ablation} report the full 
results, which exhibit consistent trends on both benchmarks.
At short horizons the models behave similarly: on MetaWorld, probe 
errors at $t{+}4$ are nearly identical across all variants.
The differences emerge at longer horizons: on RoboTwin~2.0 at 
$t{+}48$, AdaPolicy reaches $9.44$ px, while future-feature 
supervision and the baseline degrade to $16.11$ px and $17.62$ px, 
respectively; on MetaWorld at $t{+}16$, AdaPolicy attains $3.24$ px 
against $3.87$ px (baseline) and $3.61$ px (future-feature 
supervision).
Averaged over horizons, the evolved state is more linearly decodable 
than the condition state for AdaPolicy ($9.45$ vs.\ $10.20$ px on 
RoboTwin~2.0), while the two sources remain close for the baseline 
($15.76$ vs.\ $16.38$ px).
This gap indicates that the next-chunk gradient---rather than the 
architecture alone---is what organizes future information into a 
linearly accessible format.
Training with explicit future-feature targets narrows part of the 
gap to the baseline but still trails AdaPolicy on both benchmarks 
($2.55$ vs.\ $2.48$ px on MetaWorld; $14.07$ vs.\ $9.45$ px on 
RoboTwin~2.0), suggesting that observation-derived targets are only 
indirectly aligned with control.
Lastly, replacing the trainable unrolled head with its EMA-updated 
counterpart reduces the average probe error on MetaWorld from 
$2.61$ px to $2.48$ px, in line with the sensitivity results in 
Table~\ref{tab:metaworld_ablation}.

\section{Evaluation Settings}
\label{app:eval_setting}

\begin{table}[h]
\centering
\caption{Embodiment, action space, and camera setup across the four evaluation settings.}
\label{tab:env_setup}
\small
\begin{tabular}{@{}llll@{}}
\toprule
Setting & Embodiment & Action space & Camera views \\
\midrule
RoboTwin 2.0 & \begin{tabular}[c]{@{}l@{}}ALOHA-AgileX\\bimanual\end{tabular} 
  & \begin{tabular}[c]{@{}l@{}}14-D absolute joint + gripper\\(7-D per arm: 6 joints + 1 gripper)\end{tabular} 
  & \begin{tabular}[c]{@{}l@{}}one fixed third-person\\(head) camera\end{tabular} \\
\addlinespace
MetaWorld & Sawyer single arm 
  & \begin{tabular}[c]{@{}l@{}}4-D end-effector displacement\\+ gripper\end{tabular} 
  & \begin{tabular}[c]{@{}l@{}}two third-person cameras\\(corner, behindGripper)\end{tabular} \\
\addlinespace
LIBERO & Franka Panda single arm 
  & \begin{tabular}[c]{@{}l@{}}7-D operational-space delta\\+ gripper\end{tabular} 
  & wrist camera only \\
\addlinespace
Real world & \begin{tabular}[c]{@{}l@{}}AgileX Piper\\single arm\end{tabular} 
  & \begin{tabular}[c]{@{}l@{}}7-D absolute joint + gripper\\(6 joints + 1 gripper)\end{tabular} 
  & \begin{tabular}[c]{@{}l@{}}two fixed third-person\\RealSense D435 cameras\end{tabular} \\
\bottomrule
\end{tabular}
\end{table}

The four settings span diverse embodiments, action spaces, and 
camera configurations (Table~\ref{tab:env_setup}). 
RoboTwin 2.0 and the real-world setup use absolute joint-space commands 
($14$-D bimanual and $7$-D single-arm, respectively), while MetaWorld and 
LIBERO adopt end-effector-space actions with different camera configurations; 
in particular, LIBERO is evaluated with only the wrist camera. 
We use the same model architecture and hyperparameters across all settings, 
adjusting only embodiment-dependent dimensions such as the action dimension. 
The same next-chunk prediction objective is thus evaluated for bimanual 
coordination (RoboTwin 2.0), multi-task end-effector control (MetaWorld), 
wrist-view manipulation (LIBERO), and single-arm precision tasks (real world), 
demonstrating that the objective is agnostic to the action parameterization.

\begin{table}[h]
\centering
\caption{Implementation details of AdaPolicy.
Settings shared across benchmarks are shown as single entries.}
\label{tab:impl_details}
\small
\begin{tabular}{llcc}
\toprule
Category & Setting & MetaWorld \& LIBERO & RoboTwin 2.0 \\
\midrule
Input & Image resolution & $256{\times}256$ & $320{\times}240$ \\
\midrule
\multirow{3}{*}{Architecture}
 & Feature dimension & \multicolumn{2}{c}{512} \\
 & Adapter query number & 32 & 64 \\
 & Adapter layers & \multicolumn{2}{c}{4} \\
\midrule
\multirow{7}{*}{Optimization}
 & Optimizer & \multicolumn{2}{c}{AdamW} \\
 & Learning rate & \multicolumn{2}{c}{$2{\times}10^{-4}$} \\
 & Minimum learning rate & \multicolumn{2}{c}{$5{\times}10^{-5}$} \\
 & Weight decay & \multicolumn{2}{c}{$1{\times}10^{-2}$} \\
 & Betas & \multicolumn{2}{c}{$(0.9,\ 0.99)$} \\
 & Gradient clip norm & \multicolumn{2}{c}{1.0} \\
 & Batch size & \multicolumn{2}{c}{128} \\
\midrule
\multirow{3}{*}{Flow matching}
 & Timestep sampler & \multicolumn{2}{c}{logit-normal} \\
 & Sampler mean $\mu$ & \multicolumn{2}{c}{0.0} \\
 & Sampler std.\ $\sigma$ & \multicolumn{2}{c}{1.0} \\
\bottomrule
\end{tabular}
\end{table}

\paragraph{Future-feature supervision baseline.}
For the future-feature supervision variant, we extract the 
supervision target from the observation at $t{+}h$, i.e., the 
last frame covered by the current action chunk, using a frozen 
DINOv3 encoder, and take its intermediate-layer features as the 
alignment target.
A lightweight decoder follows the Q-Former 
architecture with 4 layers, each consisting of 
self-attention and cross-attention: the cross-attention queries 
are a set of learnable tokens, while the keys and values are the 
evolved state produced by the Online Action Head.
The decoder output is aligned with the future features using a 
cosine loss with a default weight of $0.5$.

\section{Gradient Stability Analysis of EMA Parameterization}
\label{app:ema_gradient}

In Section~\ref{sec:experiments} (Table~\ref{tab:metaworld_ablation}), we observed that parameterizing the Unrolled Action Head $H_{\bar\theta}$ as an exponential moving average (EMA) of the Online Action Head $H_\theta$ outperforms directly reusing a gradient-detached copy of the online parameters. 
To empirically verify whether the slowly evolving EMA target stabilizes the supervisory signal, we track the backpropagation dynamics of the next-chunk objective $\mathcal{L}_{\mathrm{next}}$ during training. 
Specifically, we compare the full \textbf{AdaPolicy} against the \emph{w/o EMA} variant (which directly uses the detached online head for the second pass) under identical training steps and data batches, recording the $\ell_2$ norm mean, standard deviation, and variance of the next-chunk gradients flowing back into the online pathway.

As reported in Table~\ref{tab:app_ema_grad}, removing EMA leads to a larger gradient magnitude accompanied by severe fluctuations, with a gradient norm variance of $0.01300$. 
Because the online parameters shift at every optimization step, directly reusing them in the unrolled head creates a rapidly moving target, which injects high-variance gradients into the evolved latent state $\hat{\mathbf{z}}_{t+h}$. 
In contrast, maintaining a slowly evolving EMA unrolled head reduces the gradient norm variance by nearly half. 
This suggests that the EMA parameterization provides a stable, lower-variance shaping signal for learning anticipatory representations.

\begin{table}[h]
\centering
\caption{
\textbf{Comparison of next-chunk gradient statistics with and without EMA.} 
Maintaining an EMA-updated unrolled head substantially reduces the variance of the gradients backpropagated from the next-chunk loss $\mathcal{L}_{\mathrm{next}}$.
}
\label{tab:app_ema_grad}
\small
\setlength{\tabcolsep}{10pt}
\begin{tabular}{lcc}
\toprule
Unrolled Head Variant & Next-Chunk Gradient Norm (Mean $\pm$ Std) & Variance \\
\midrule
w/o EMA (detached online copy) & $0.3430 \pm 0.1140$ & $0.01300$ \\
\textbf{w/ EMA (AdaPolicy)}    & $\mathbf{0.2818 \pm 0.0829}$ & $\mathbf{0.00688}$ \\
\bottomrule
\end{tabular}
\end{table}

\section{Additional Planning Probe on Condition States ($c_t$)}
\label{app:cem_cond}

\paragraph{Overview and results.}
While the latent-space planning experiment in the main paper directly queries the Online Action Head using the evolved state ($\hat{z}_{t+h}$), we conduct an additional probe here to examine whether the upstream \emph{condition state} ($c_t$, output by the Context Adapter prior to the action head) also benefits from next-chunk supervision.
Following the evaluation protocol of LeWorldModel~\cite{lewm}, we freeze the policy's visual encoder and Context Adapter, train a separate lightweight dynamics predictor on top of the extracted $c_t$ tokens for each task, and perform CEM trajectory selection toward a goal observation $20$ environment steps ahead.

As shown in Table~\ref{tab:cem_cond}, planning over AdaPolicy's condition state reaches the goal (within $5$\,cm) at $23.1\%$ on average across seven MetaWorld tasks, improving over the baseline's feature space ($17.4\%$).
While performance is comparable on tasks with subtle object motions (and slightly lower on \emph{basketball}), the gains are clear on tasks involving larger object displacements, such as \emph{stick-push} and \emph{pick-out-of-hole}.
Consistent with our ablation on adapter gradients (Table~5), these results indicate that backpropagating next-chunk errors into the Context Adapter helps regularize the upstream conditioning tokens $c_t$, making them more amenable to downstream dynamics fitting even before entering the action head.

\paragraph{Predictor architecture and training.}
Each observation frame is encoded into $32$ conditioning tokens of dimension $512$ using the frozen Context Adapter (identical to the feature-extraction pipeline in our linear probing experiments).
The goal representation is the cached $c_t$ feature of the observation $20$ environment steps ahead from the same test episode.
On top of the frozen $c_t$ tokens, we train a task-specific latent dynamics predictor: a $4$-layer cross-attention transformer that takes the current conditioning tokens and a $5$-step action block as input to predict the future conditioning tokens, supervised by an MSE loss.
For each task, episodes $0$--$24$ of the $50$ demonstrations are used for training, episodes $25$--$29$ for validation (selecting the best validation checkpoint), and episodes $30$--$49$ are held out for evaluation.
The predictor is optimized using AdamW (learning rate $3{\times}10^{-4}$, weight decay $10^{-3}$) with a batch size of $256$ for $10{,}000$ steps.

\begin{table}[h]
\centering
\caption{
CEM planning over frozen condition states ($c_t$) on seven MetaWorld tasks.
A separate latent dynamics predictor is trained on top of the frozen $c_t$ tokens to plan toward a goal $20$ environment steps ahead.
}
\label{tab:cem_cond}
\small
\setlength{\tabcolsep}{4pt}
\begin{tabular}{lcccccccc}
\toprule
 & \rotatebox{0}{lever-pull} & basketball & peg-insert-side & stick-push & pick-out-of-hole & sweep & shelf-place & Avg. \\
\midrule
Baseline  & 54 & \textbf{34} & 14 & 10 & 2  & 4  & 4 & 17.4 \\
AdaPolicy & \textbf{64} & 24 & \textbf{16} & \textbf{26} & \textbf{16} & \textbf{12} & 4 & \textbf{23.1} \\
\bottomrule
\end{tabular}
\end{table}

\paragraph{CEM search protocol.}
Following~\cite{lewm}, actions are grouped into blocks of $5$ environment steps.
At each planning step, CEM samples $300$ candidate action sequences over a horizon of $4$ blocks by unrolling the learned predictor autoregressively.
Candidates are scored by the squared $L_2$ distance between the predicted terminal tokens and the goal tokens, and the top-$30$ elites are used to update the Gaussian sampling distribution over $30$ iterations.
We execute only the first $5$-step action block in the environment and replan from the newly encoded observation.
We evaluate $50$ test instances per task drawn from the held-out episodes, requiring an initial object displacement above $5$\,cm within the goal window to exclude trivial stationary cases.
An instance is counted as successful if the target object comes within $5$\,cm of the privileged ground-truth goal position at any step within the execution budget (reach@$5$\,cm).

\begin{table*}[h]
\centering
\setlength{\tabcolsep}{2.5pt}
\renewcommand{\arraystretch}{0.95}
\scriptsize

\resizebox{\textwidth}{!}{%
\begin{tabular}{lcccc lcccc lcccc}
\toprule
& \multicolumn{2}{c}{X-VLA} & \multicolumn{2}{c}{AdaPolicy}
& & \multicolumn{2}{c}{X-VLA} & \multicolumn{2}{c}{AdaPolicy}
& & \multicolumn{2}{c}{X-VLA} & \multicolumn{2}{c}{AdaPolicy} \\
\cmidrule(lr){2-3} \cmidrule(lr){4-5}
\cmidrule(lr){7-8} \cmidrule(lr){9-10}
\cmidrule(lr){12-13} \cmidrule(lr){14-15}
Task & Clean & Rand. & Clean & Rand.
& Task & Clean & Rand. & Clean & Rand.
& Task & Clean & Rand. & Clean & Rand. \\
\midrule

Adjust Bottle
& 100 & 99 & 100 & 100
& Open Microwave
& 79 & 71 & 82 & 58
& Place Object Stand
& 86 & 88 & 84 & 94 \\

Beat Block Hammer
& 92 & 88 & 94 & 80
& Pick Diverse Bottles
& 58 & 36 & 72 & 84
& Place Phone Stand
& 88 & 87 & 82 & 88 \\

Blocks Ranking RGB
& 83 & 83 & 100 & 96
& Pick Dual Bottles
& 47 & 36 & 92 & 92
& Place Shoe
& 96 & 95 & 100 & 100 \\

Blocks Ranking Size
& 67 & 74 & 62 & 74
& Place A2B Left
& 48 & 49 & 58 & 76
& Press Stapler
& 92 & 98 & 98 & 96 \\

Click Alarmclock
& 99 & 99 & 92 & 88
& Place A2B Right
& 36 & 36 & 56 & 54
& Put Bottles Dustbin
& 74 & 77 & 90 & 92 \\

Click Bell
& 100 & 100 & 96 & 90
& Place Bread Basket
& 81 & 71 & 90 & 88
& Put Object Cabinet
& 46 & 48 & 86 & 86 \\

Dump Bin Bigbin
& 79 & 77 & 92 & 90
& Place Bread Skillet
& 77 & 67 & 88 & 88
& Rotate QRcode
& 34 & 33 & 92 & 88 \\

Grab Roller
& 100 & 100 & 100 & 100
& Place Burger Fries
& 94 & 94 & 96 & 96
& Scan Object
& 14 & 36 & 78 & 82 \\

Handover Block
& 73 & 37 & 88 & 62
& Place Can Basket
& 49 & 52 & 68 & 76
& Shake Bottle
& 99 & 100 & 100 & 100 \\

Handover Mic
& 0 & 0 & 92 & 98
& Place Cans Plasticbox
& 97 & 98 & 90 & 88
& Shake Bottle Horizontally
& 100 & 100 & 100 & 98 \\

Hanging Mug
& 23 & 27 & 50 & 46
& Place Container Plate
& 97 & 95 & 98 & 100
& Stack Blocks Three
& 6 & 10 & 70 & 76 \\

Lift Pot
& 99 & 100 & 98 & 96
& Place Dual Shoes
& 79 & 88 & 66 & 46
& Stack Blocks Two
& 92 & 87 & 94 & 94 \\

Move Can Pot
& 89 & 86 & 84 & 92
& Place Empty Cup
& 100 & 98 & 98 & 98
& Stack Bowls Three
& 76 & 86 & 80 & 86 \\

Move Pillbottle Pad
& 73 & 71 & 94 & 74
& Place Fan
& 80 & 75 & 86 & 92
& Stack Bowls Two
& 96 & 93 & 90 & 96 \\

Move Playingcard Away
& 93 & 98 & 96 & 96
& Place Mouse Pad
& 70 & 70 & 62 & 50
& Stamp Seal
& 76 & 82 & 44 & 40 \\

Move Stapler Pad
& 78 & 73 & 70 & 70
& Place Object Basket
& 44 & 39 & 78 & 88
& Turn Switch
& 40 & 61 & 78 & 78 \\

Open Laptop
& 93 & 100 & 98 & 98
& Place Object Scale
& 52 & 74 & 94 & 84
& \multicolumn{4}{c}{} \\

\midrule
\multicolumn{15}{c}{%
\textbf{Average:}\quad
X-VLA --- Clean: 72.88, Rand.: 72.84;\qquad
\textbf{AdaPolicy --- Clean: 84.92, Rand.: 84.04}} \\
\bottomrule
\end{tabular}%
}

\caption{Per-task success rates on 50 RoboTwin~2.0 tasks under clean
and randomized settings.}
\label{tab:adapolicy_robotwin_per_task}
\end{table*}

\end{document}